\documentclass[letterpaper, 10 pt, journal]{IEEEtran}

\usepackage{graphics}
\usepackage{times}
\usepackage{amsmath}            
\usepackage{amssymb}            
\usepackage{graphicx}
\usepackage{algorithm}
\usepackage[noend]{algpseudocode}
\usepackage[noend]{algpseudocode}
\usepackage{booktabs}
\usepackage{color}
\usepackage{multirow}
\usepackage{subcaption}
\usepackage{rotating}
\usepackage{cite}
\usepackage{soul}
\usepackage{pifont}
\usepackage{url}
\usepackage{xcolor, colortbl}

\definecolor{instructioncolor}{rgb}{.5,.5,.5}
\definecolor{lightgray}{RGB}{210, 210, 210}
\definecolor{darkred}{rgb}{.804,.196,.196}
\definecolor{darkorange}{rgb}{1.0,.55,.0}
\definecolor{darkgreen}{rgb}{.196,.70,.196}
\definecolor{table1cellcolor}{rgb}{1.0,1.0,0.6}

\sethlcolor{lightgray}

\newcommand{\xmark}{\ding{55}}

\newcommand{\threedgreen}{\textcolor{darkgreen}{3D}}
\newcommand{\twodred}{\textcolor{darkred}{2D}}
\newcommand{\cmarkgreen}{\textcolor{darkgreen}{\ding{51}}}
\newcommand{\xmarkred}{\textcolor{darkred}{\ding{55}}}
\newcommand{\trimark}{\textcolor{darkorange}{$\boldsymbol{\triangle}$}}

\definecolor{bestcolor}{rgb}{.196,.70,.196}   \definecolor{secondcolor}{rgb}{.096,.096,.9} \newcommand{\best}[1]{\textcolor{bestcolor}{\textbf{#1}}}
\newcommand{\second}[1]{\textcolor{secondcolor}{\textbf{#1}}}

\def\secref#1{Section~\ref{#1}}
\def\figref#1{Fig.~\ref{#1}}
\def\tabref#1{Table~\ref{#1}}
\def\eqref#1{(\ref{#1})}

\def\vsfig{\vspace{-0.3cm}}
\def\vstab{\vspace{-0.3cm}}
\def\vsequ{\vspace{-0.15cm}}
\def\vseq{\vspace{-0.3cm}}

\newcommand{\rom}[1]{\uppercase\expandafter{\romannumeral #1\relax}}

\makeatletter
\usepackage{xspace}
\DeclareRobustCommand\onedot{\futurelet\@let@token\@onedot}
\def\@onedot{\ifx\@let@token.\else.\null\fi\xspace}
 
\def\ie{i.e\onedot}

\def\etal{{\textit{et al}}\onedot}
\makeatother

\def\etalcite#1{\etal~\cite{#1}}

\usepackage{array}
\newcolumntype{L}[1]{>{\raggedright\let\newline\\\arraybackslash\hspace{0pt}}m{#1}}
\newcolumntype{C}[1]{>{\centering\let\newline\\\arraybackslash\hspace{0pt}}m{#1}}
\newcolumntype{R}[1]{>{\raggedleft\let\newline\\\arraybackslash\hspace{0pt}}m{#1}}

\def\argmin{\mathop{\rm argmin}}
\def\min{\mathop{\rm min}}
\def\max{\mathop{\rm max}}

\newcommand{\state}{\mathbf{x}}

\title{\LARGE \bf Multi-Mapcher: Loop Closure Detection-Free Heterogeneous LiDAR Multi-Session SLAM Leveraging Outlier-Robust Registration \\ for Autonomous Vehicles}

\author{Hyungtae Lim$^{1\dagger}$, Daebeom Kim$^{2\dagger}$, and Hyun Myung$^{2*}$, \textit{Senior Member, IEEE}\thanks{$^*$Corresponding author: Hyun Myung}
  \thanks{$^\dagger$Both authors have equally contributed.}
  \thanks{$^{1}$Hyungtae Lim is with the Laboratory for Information \& Decision Systems, Massachusetts Institute of Technology, Cambridge, MA 02139, USA. {\tt\scriptsize shapelim@mit.edu}}
  \thanks{$^{2}$Daebeom Kim ahd Hyun Myung are with the School of Electrical Engineering, KAIST (Korea Advanced Institute of Science and Technology), Daejeon, 34141, Republic of Korea. {\tt\scriptsize ted97k@kaist.ac.kr, hmyung@kaist.ac.kr} \hfill \break}
}

\begin{document}
\maketitle

\begin{abstract}
  As various 3D light detection and ranging (LiDAR) sensors have been introduced to the market, research on multi-session simultaneous localization and mapping~(MSS) using heterogeneous LiDAR sensors has been actively conducted.
  Existing MSS methods mostly rely on loop closure detection for inter-session alignment; however, the performance of loop closure detection can be potentially degraded owing to the differences in the density and field of view~(FoV) of the sensors used in different sessions.
  In this study, we challenge the existing paradigm that relies heavily on loop detection modules and propose a novel MSS framework, called \textit{Multi-Mapcher}, that employs large-scale map-to-map registration to perform inter-session initial alignment, which is commonly assumed to be infeasible, by leveraging outlier-robust 3D point cloud registration.
  Next, after finding inter-session loops by radius search based on the assumption that the inter-session initial alignment is sufficiently precise, anchor node-based robust pose graph optimization is employed to build a consistent global map.
  As demonstrated in our experiments, our approach shows substantially better MSS performance for various LiDAR sensors used to capture the sessions and is faster than state-of-the-art approaches.
  Our code is available at https://github.com/url-kaist/multi-mapcher.
\end{abstract}


\section{Introduction}
\label{sec:intro}

\IEEEPARstart{M}{ulti-session} simultaneous localization and mapping~(Multi-session SLAM, or MSS) is an approach for aligning multiple maps collected by autonomous vehicles or robots over time to build a single comprehensive map in a global coordinate system~\cite{kim2022lt, lajoie2023swarm, kim2010multiple, mcdonald2013real, ozog2016long}.
Here, the term \textit{session} refers to the independent exploration or data collection activities performed by a robotic vehicle (or agent) over a specific period.
Because the MSS enables autonomous driving systems to estimate the relative poses in different sessions with respect to the same global coordinate system,
numerous researchers have proposed novel approaches for long-term map management~\cite{carpin2008fast,bonanni2014merging,bonanni20173d, kim2022road, yu2023multi, yuan2023lta, chang2022lamp} or online collaborative tasks of autonomous vehicles and swarm robots~\cite{lazaro2013multi, labbe2014online, cieslewski2018data, lajoie2020door, zhu2023swarm, huang2021disco, zhong2023dcl}.

Meanwhile, a variety of 3D light detection and ranging~(LiDAR) sensors has been introduced to the market, including mechanically spinning omnidirectional LiDAR~(Omni-LiDAR), solid-state LiDAR~(Solid-LiDAR), and flash type LiDAR sensors~\cite{lim2024helimos}.
These advancements have led to the development of diverse types of self-driving vehicles equipped with various LiDAR sensors.
In light of this consideration, MSS that can handle sessions captured with heterogeneous LiDAR sensors is necessary to achieve long-term autonomy for these diverse platforms.

\begin{figure}[t!]
    \centering
    \def\twosubfigsize{0.48\textwidth}
    \begin{subfigure}[b]{\twosubfigsize}
        \centering
        \includegraphics[width=1.0\textwidth]{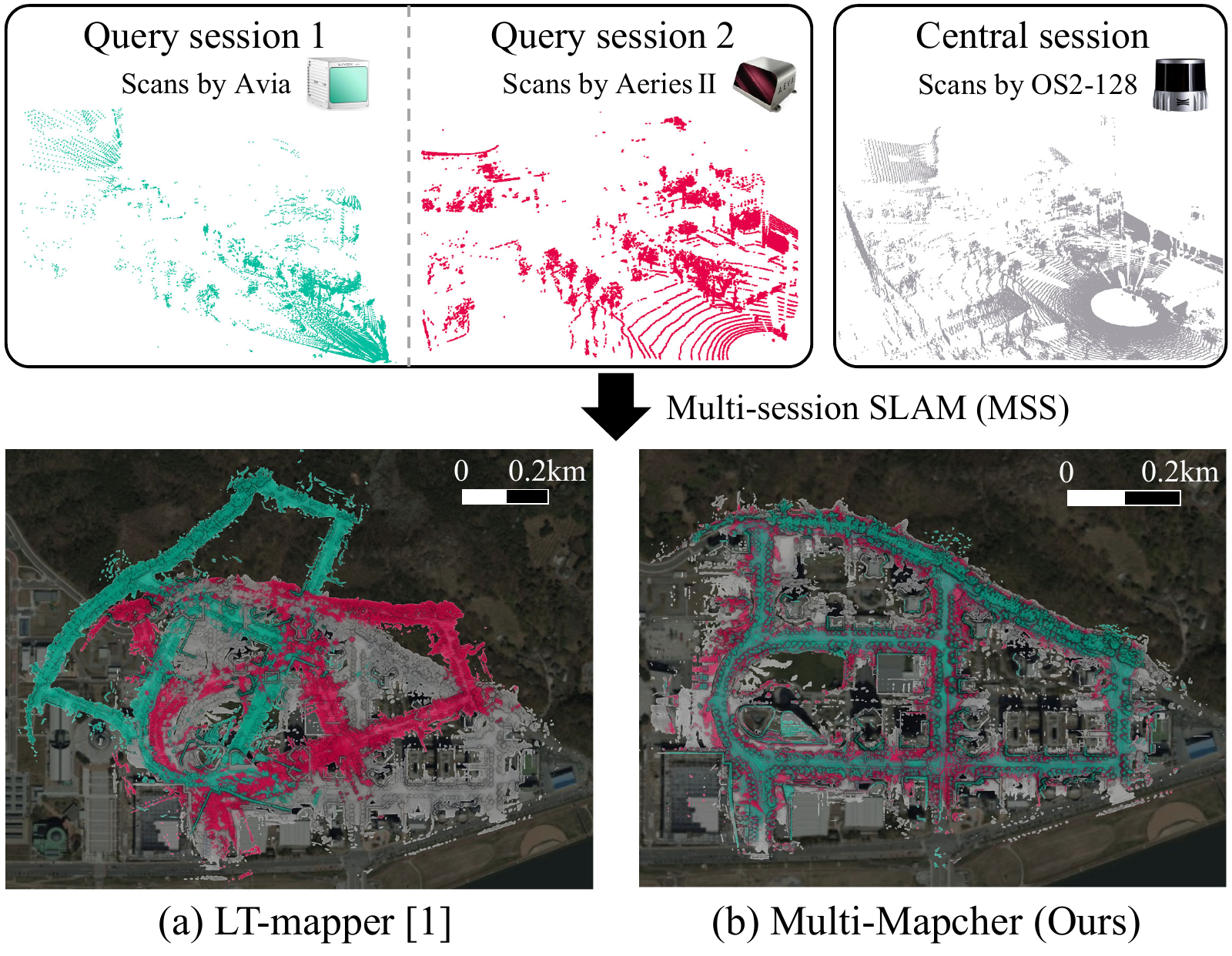}
    \end{subfigure}
    \captionsetup{font=footnotesize}
    \caption{Multi-session simultaneous localization and mapping~(MSS) results of (a)~LT-mapper~\cite{kim2022lt}, which is a baseline that relies heavily on a loop closure detection~(LCD) module when finding inter-session loop pairs, and (b)~our proposed method called, \textit{Multi-Mapcher}.
    Gray, dark cyan, and dark magenta colors indicate each session obtained by Ouster OS2-128, Livox Avia, and Aeva Aeries~\rom{2}, respectively.
    Note that our Multi-Mapcher robustly aligns different sessions from heterogeneous LiDAR sensors while minimizing the dependency on the LCD modules~(best viewed in color).}
    \label{fig:motivation}
    \vsfig
\end{figure}

\begin{table*}[t!]
    \centering
    \captionsetup{font=footnotesize}
    \caption{Comparison between existing centralized multi-session (or map merging) SLAM approaches and our approach. The term \textit{Deformation-robust} indicates whether a method works in a robust manner even if single session maps containing inherent pose errors are given.
    The symbol \trimark$\,$ indicates that a method exploits occupancy grid map representation, so it is applicable to heterogeneous LiDAR sensor setups.}
    \setlength{\tabcolsep}{4pt}
    {\scriptsize
        \begin{tabular}{l|ccccccc}
            \toprule \midrule
            Approach & Year &  2D or 3D  & \begin{tabular}{@{}c@{}}Large scale \\ ($>$ 1,000 scans and \\ $>$ several kilometers) \end{tabular} & \begin{tabular}{@{}c@{}}Deformation \\ -robust \end{tabular} & \begin{tabular}{@{}c@{}}Available when \\ the inter-session pose \\ difference is unknown \end{tabular} & \begin{tabular}{@{}c@{}}Loop closing detection (LCD) \\ -free (i.e. not affected by \\ performance degradation of \\ LCD approaches)\end{tabular} & \begin{tabular}{@{}c@{}}Robust in\\ heterogeneous \\LiDAR sensor setups\end{tabular} \\ \midrule
            Carpin~\etalcite{carpin2008fast} & 2008 & \twodred  & \xmarkred & \xmarkred  & \cmarkgreen & \cmarkgreen & \trimark \\
            Bonanni~\etalcite{bonanni2014merging}  & 2014 & \twodred  & \xmarkred & \xmarkred  & \xmarkred & \cmarkgreen  & \trimark \\
            Bonanni~\etalcite{bonanni20173d}  & 2017 & \threedgreen  & \xmarkred & \cmarkgreen  & \cmarkgreen & \xmarkred & \xmarkred \\
            Kim and Kim~\cite{kim2022lt} & 2022 & \threedgreen  & \cmarkgreen & \cmarkgreen  & \cmarkgreen & \xmarkred & \xmarkred \\
            Yuan \etalcite{yuan2023lta} & 2024 & \threedgreen  & \cmarkgreen & \cmarkgreen  & \cmarkgreen & \xmarkred & \xmarkred \\
            Stathoulopoulos~\etalcite{stathoulopoulos2024frame} & 2024 & \threedgreen  & \cmarkgreen & \cmarkgreen  & \cmarkgreen & \xmarkred & \xmarkred \\
            \cellcolor{table1cellcolor}Proposed & \cellcolor{table1cellcolor}2024 & \cellcolor{table1cellcolor}\threedgreen & \cellcolor{table1cellcolor}\cmarkgreen & \cellcolor{table1cellcolor}\cmarkgreen & \cellcolor{table1cellcolor}\cmarkgreen & \cellcolor{table1cellcolor}\cmarkgreen & \cellcolor{table1cellcolor}\cmarkgreen\\
            \midrule \bottomrule
        \end{tabular}
    }
    \label{table:summary}
\end{table*} 
However, we observed that recent MSS approaches~\cite{kim2022lt, huang2021disco, yuan2023lta,stathoulopoulos2024frame} may not perform well between two sessions acquired by heterogeneous LiDAR sensors, specifically between a session captured with an Omni-LiDAR and another with a Solid-LiDAR sensor, or vice versa~(note that, in this study, flash type LiDAR sensors are beyond our scope because of the poor resolution by now).
These approaches rely heavily on loop closure detection~(LCD) modules to establish associations between the same locations in different sessions, which are referred to as inter-session loops.
Unfortunately, as Jung~\etalcite{jung2023helipr} reported, the performance of the existing LCD modules used in MSS methods~\cite{kim2018scancontext,vid2022logg3d, xu2023ring++} is substantially degraded when a 3D point cloud from a different type of LiDAR sensor is taken as a query.
Consequently, as shown in \figref{fig:motivation}(a), the failure to find inter-session loops leads to a misalignment between sessions, resulting in a distorted and inconsistent point cloud map.

Therefore, we challenge the existing paradigm that relies on LCD modules for inter-session alignment.
To this end, we address the challenge of initial alignment through a large-scale map-to-map registration, which is commonly assumed as infeasible~\cite{bonanni20173d}, by leveraging outlier-robust point cloud registration~\cite{lim2023quatro++}.
In particular, this outlier-robust registration is employed not only at the map-to-map level but also at the scan-to-scan level to enhance inter-session loop closing.
Finally, anchor node-based pose graph optimization~(PGO) for multiple sessions~\cite{kim2010multiple, mcdonald2013real, ozog2016long} is performed to refine the alignment and build a consistent global map, as shown in~\figref{fig:motivation}(b).

The main contribution of this study is a novel LCD-free MSS approach, called \textit{Multi-Mapcher}, which is a combination of the words \textit{Map} and \textit{Matcher} for \textit{multi}-session SLAM.
To the best of our knowledge, this is the first study that extends the concept of map merging, which is usually used for 2D occupancy grid maps~\cite{carpin2008fast}, to 3D space with different types of LiDAR sensors.
Specifically, we demonstrate that inter-session initial alignment in 3D space is possible through outlier-robust registration, even when the map of each session is somewhat imprecise or deformed.

In summary, we make the following three key claims.
By exploiting map-to-map and scan-to-scan level outlier-robust registration, our LCD-free approach (i)~precisely builds a consistent global map even when sessions with different types of LiDAR sensors are provided,
(ii)~shows robustness against both low- and high-dynamic changes in the surroundings, as well as in partially overlapped sessions,
and (iii)~is faster and more efficient than existing scan-to-scan LCD-based MSS approaches.
These claims are backed up by the following sections and by our experimental evaluation.

\section{Related Work}
\label{sec:related}

Over the past few years, numerous studies have significantly advanced LiDAR SLAM, focusing primarily on the development of various methods for accurate mapping in single session SLAM~\cite{wang2020intelligent,gkillas2023federated,tian2023dl,xia2023integrated,zhao2019lidar,kim2024awvmoslio,he2024igicp,zhou2023hpplo,lim2023ur,jo2024real,zhang2022intelligent,liu2023software}.
However, despite these advancements, there remains a persistent demand for map merging or MSS techniques to achieve long-term map update because these approaches are essential for managing map data acquired over different time steps~\cite{berrio2021long,ma2024mmlins,liang2022hierarchical}.

In general, MSS is mainly classified into two groups:
a)~online and distributed MSS for the relative pose estimation of multiple robots~\cite{labbe2014online, zhu2023swarm, huang2021disco, zhong2023dcl, tian2022kimera}
and b)~centralized MSS, which focuses on long-term map management.
In this paper, we place more emphasis on centralized MSS and map merging for long-term map updates.
To help the reader's understanding, we summarize the differences between our approach and state-of-the-art centralized MSS approaches in \tabref{table:summary}.

Unlike centralized MSS, decentralized MSS accounts for limited communication, that is, narrow bandwidth, limited communication range, and the computational resources of robots.
Kim~\etalcite{kim2010multiple} first introduced the concept of an anchor node to address the under-constrained trajectory problem in each session when optimizing multiple pose graphs.
Labbe and Michaud~\cite{labbe2014online} demonstrated the feasibility of decentralized MSS in a large-scale scenario by employing online global loop closure detection and efficient memory management.
Lajoie~\etalcite{lajoie2020door} proposed DOOR-SLAM, which is based on peer-to-peer communication and first proposed an outlier-robust pairwise consistency maximization~(PCM) algorithm to reject spurious inter-session loops.
As an extension of DOOR-SLAM, Lajoie~\etalcite{lajoie2023swarm} proposed Swarm-SLAM, which supports multi-modal sensor suites and employs inter-session loop closure prioritization to accelerate convergence and reduce communication overhead.

About ten years ago, the centralized MSS research primarily focused on accurately aligning occupancy grid maps in 2D.
For instance, Carpin~\cite{carpin2008fast} proposed a fast and accurate Hough transform-based map merging algorithm for 2D occupancy grid maps and this approach was extended to MSS for multiple robots by Saeedi~\etalcite{saeedi2014map}.
Bonanni~\etalcite{bonanni2014merging} proposed a novel Voronoi diagram-based consistent 2D map merging approach, yet this approach strongly assumed that the initial alignment was sufficiently accurate.
To relax this assumption and extend the approach to 3D maps, Bonanni~\etalcite{bonanni20173d} proposed the first consistent 3D MSS approach by exploiting 3D point cloud data.

In recent years, LAMP\;2.0~\cite{chang2022lamp} and Maplab\;2.0~\cite{cramariuc2022maplab} have become renowned centralized MSS systems that are designed to manage large-scale maps and heterogeneous robot configurations.
Chang~\etalcite{chang2022lamp} demonstrated that LAMP\;2.0 can build a consistent map in tunnel-like degenerate scenes by employing an outlier-resilient PGO based on graduated non-convexity~(GNC)~\cite{yang2020teaser}.
Cramariuc~\etalcite{cramariuc2022maplab} proposed Maplab 2.0 to easily integrate data from multi-modal sensor setups, deep learning-based features, and LCD modules.
Yuan~\etalcite{yuan2023lta} proposed LTA-OM, which exploits the stable triangle descriptor~(STD)~\cite{yuan2023std} to perform LCD and suggested a novel factor marginalization.
Beyond building a consistent global map, Kim and Kim~\cite{kim2022lt}, and Yang~\etalcite{yang2024lifelong} demonstrated the importance of incorporating dynamic point removal and environmental change detection modules across different sessions.

Existing approaches present front-end agnostic characteristics~\cite{kim2022lt}, that is, they are independent of LiDAR odometry algorithms and are applicable to heterogeneous robot setups~\cite{chang2022lamp, lajoie2023swarm}. However, to the best of our knowledge, few studies have been designed to handle heterogeneous LiDAR sensor setups.
Certainly, this issue can be addressed through sensor-agnostic representations such as occupancy grid maps~\cite{carpin2008fast, bonanni2014merging}, but these occupancy grid-based approaches might be difficult to apply in large-scale environments where the values of $z$, roll, and pitch in the trajectory change.
Furthermore, as described above, existing approaches mainly depend on LCD modules to search the inter-session loops for the inter-session initial alignment.
However, some LCD modules tend to generate inconsistent descriptors, even for data measured within the same scene from heterogeneous LiDAR sensors,
highlighting the difficulty of maintaining descriptor consistency across different sensor types~\cite{jung2023helipr}.
Consequently, this limitation leads to erroneous inter-session loop detection and, ultimately, catastrophic failure in the inter-session alignment.

Therefore, instead of relying on LCD modules, we accumulate point cloud data from each heterogeneous LiDAR sensor to build point cloud maps and perform an initial alignment through large-scale map-to-map registration.
Thus, the initial alignment is less affected by differences in the sparsity or measurement patterns of each scan.
Our initial alignment is similar to that of Hydra-Multi~\cite{chang2023hydra}, which also has a robust distributed initialization using registration between scene graph nodes from each session.
However, our approach differs from Hydra-Multi in that ours does not use additional semantic representations and focuses on MSS between sessions from heterogeneous LiDAR sensor setups.
 
\section{Multi-Mapcher: Our Approach to Multi-Session SLAM}
\label{sec:main}

\begin{figure*}[t!]
    \centering
    \def\twosubfigsize{0.95\textwidth}
    \begin{subfigure}[b]{\twosubfigsize}
        \centering
        \includegraphics[width=1.0\textwidth]{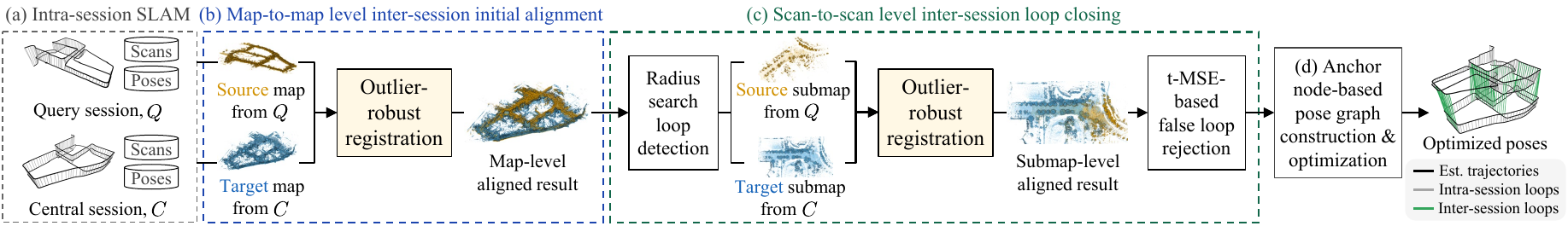}
    \end{subfigure}
    \captionsetup{font=footnotesize}
    \caption{Overview of our multi-session SLAM framework, called \textit{Multi-Mapcher}, which consists of four steps~(best viewed in color). (a)~First, intra-session SLAM is performed using existing SLAM frameworks~\cite{wei2022fastlio}, which outputs scans and optimized poses, including intra-session odometry and loop constraints. (b) Second, map-to-map level registration is performed to initially estimate the relative pose between the reference frames of query session $Q$ and central session $C$. (c) Third, after the initial alignment between two sessions, inter-session loops can be easily detected by a radius search, which is followed by our truncated mean squared error~(t-MSE)-based false loop rejection to filter erroneous loop candidates~(see \secref{sec:scan_to_scan}). Note that our Multi-Mapcher uses the same outlier-robust registration module when performing registration at both the map-to-map and scan-to-scan levels. (d) Finally, taking all the odometry, intra-session, and inter-session constraints as inputs, anchor node-based pose graph optimization is performed to build a global map across multiple sessions.}
    \label{fig:overview}
    \vsfig
\end{figure*}

A schematic diagram of our Multi-Mapcher is shown in \figref{fig:overview}.
Our approach aims to achieve a robust and efficient multi-session SLAM by addressing the inter-session initial alignment problem using outlier-robust registration~\cite{lim2022single, lim2023quatro++}.
One remarkable aspect is that our Multi-Mapcher uses the outlier-robust registration module at the map-to-map level for the initial inter-session alignment~(\figref{fig:overview}(b)) and also at the scan-to-scan level~(\figref{fig:overview}(c)).
Consequently, even though sessions captured by heterogeneous LiDAR sensors are given as inputs, our Multi-Mapcher can successfully align them into a single global map.

\subsection{Differences in Pose Graph Optimization Between \\ Single Session and Multi-Session SLAM}

To perform PGO, a factor graph should be constructed by expressing the poses of the robot as nodes and the estimated relative pose differences as constraints.
The PGO for multi-session SLAM has two major differences compared with the single session PGO.
First, because the reference frame for the trajectory of each session itself is not fixed, expressing multiple trajectories in the same coordinate system is needed to resolve the under-constrained issue.
Second, in contrast to PGO for a single session SLAM, which only consists of odometry and loop constraints within the intra-session, PGO for multi-session SLAM requires inter-session loop constraints to align different sessions with each other and thus to share a common reference frame (or global coordinate system).

\subsection{Anchor Node-Based Pose Graph Optimization}

To address the first issue, we introduce the concept of an anchor node, which is a state that expresses the transformation matrix from the global coordinate system to the reference frame of each session~\cite{kim2010multiple, mcdonald2013real, ozog2016long}.
By doing so, it eliminates the under-constrained problem by anchoring the entire trajectory to a known location, enabling consistent and accurate alignment between sessions.

Formally, let us define the existing session as the central session~$C$ and the session to be aligned as the query session~$Q$ and their anchor nodes as $\Delta_C$ and $\Delta_Q$, respectively~\cite{kim2022lt}. Then the anchor node-based inter-session loop factor $\phi(\cdot)$ can be expressed as follows:

\begin{equation}
    \label{eqn:anchor_node_factor}
    \scalebox{0.94}{
        $\begin{aligned}
             & \phi\left(\mathbf{x}_{C, j}, \mathbf{x}_{Q, k}, \Delta_C, \Delta_Q\right) \\
             & \propto \exp \left(-\frac{1}{2}\left\|\left(\left(\Delta_C \oplus \mathbf{x}_{C, j}\right) \ominus\left(\Delta_Q \oplus \mathbf{x}_{Q, k}\right)\right)-\mathbf{z}^\text{inter}_{j,k}\right\|_{\Sigma^\text{inter}_{j,k}}^2\right)
        \end{aligned}$
    }
\end{equation}

\noindent where the form of $\|\mathbf{e}\|_{\Sigma}^2=\mathbf{e}^T \Sigma^{-1} \mathbf{e}$ is the squared Mahalanobis distance with the error vector $\mathbf{e}$ and covariance matrix $\Sigma$;
$j$ and $k$ are the node indices of the central session and query session, respectively, such that $\state_{s, i}$ is the $i$-th node of the session $s \in \{C, Q\}$ in a graph structure;
$\oplus$ and $\ominus$ are SE(3) pose transformation operators that add and subtract poses corresponding to the nodes, respectively;
$\mathbf{z}^\text{inter}_{j,k}$ is the inter-session loop measurement with its corresponding covariance~$\Sigma^\text{inter}_{j,k}$.

Therefore, based on \eqref{eqn:anchor_node_factor}, the objective function for multi-session PGO to optimize a set of all the states $\mathcal{X}$ can be formulated as follows:

\vsequ
\newcommand{\priorfactor}{\mathbf{p}}
\newcommand{\largespace}{\qquad \qquad}

\begin{equation}
    \scalebox{0.92}{
    $\begin{aligned}
        &\hat{\mathcal{X}}=
        \underset{\mathcal{X}}{\operatorname{argmin}}
        \Biggl\{ \underbrace{\left\|\priorfactor_C-\mathbf{x}_{C,0}\right\|_{\Sigma^\text{prior}}^2}_\text{Prior factor} \\
        & + \sum_{s\in\{Q, C\}} \biggl(  \underbrace{\sum_{i}\left\| \left(\state_{s,i-1} \ominus \state_{s,i} \right) - \mathbf{z}^\text{odom}_{i}\right\|_{\Sigma^\text{odom}_{i}}^{2}}_\text{Intra-session odometry constraints}\\
        & \largespace + \underbrace{\sum_{(l, m) \in \mathcal{L}^\text{intra}_s} \left\| \rho \Big( \left(\state_{s,l} \ominus \state_{s,m}\right) - \mathbf{z}^\text{intra}_{l,m}\Big)\right\|_{\Sigma^\text{intra}_{l,m} }^{2}}_\text{Intra-session loop constraints}  \biggr)\\
        & + \underbrace{\sum_{(j, k) \in \mathcal{L}^\text{inter}}\left\| \rho \Big(\left(\Delta_C \oplus \mathbf{x}_{C, j}\right) \ominus\left(\Delta_Q \oplus \mathbf{x}_{Q, k}\right))-\mathbf{z}^\text{inter}_{j,k} \Big) \right\|_{\Sigma^\text{inter}_{j,k} }^{2}}_\text{Inter-session loop constraints} \Biggl\}
    \end{aligned}$
    }
    \label{eqn:multi_session}
\end{equation}

\noindent where $\priorfactor_C$ and $\Sigma^\text{prior}$ are the state prior of the nodes of the central session and its covariance, respectively, and $\mathbf{x}_{C,0}$ is the node for the prior factor of the central session;
$\mathbf{z}^\text{odom}_{i}$ and $\mathbf{z}^\text{intra}_{l,m}$ denote the intra-session odometry and loop measurements, respectively, with their corresponding covariances $\Sigma^\text{odom}_{i}$ and $\Sigma^\text{intra}_{l,m}$;
$\mathcal{L}^\text{intra}_s$ and $\mathcal{L}^\text{inter}$ are the intra- and inter-session loops found by loop closing, respectively;
$\rho(\cdot)$ represents a robust kernel function designed to suppress the effects of erroneous loop constraints.

Note that we set the covariance of $\Delta_Q$ to be significantly larger than that of $\Delta_C$ when solving~\eqref{eqn:multi_session}.
By doing so, the transformation matrix of $\Delta_Q$ undergoes more substantial changes to minimize the error of the inter-session loop constraints, which is equivalent to aligning the query session to the central session.
Consequently, the result of \eqref{eqn:multi_session} is the optimized poses of all sessions and transformation matrices of $\Delta_C$ and $\Delta_Q$.

\newcommand{\voxel}{\nu}
\newcommand{\vmap}{\nu_{m}}
\newcommand{\vsubmap}{\nu_{s}}
\subsection{Overview of a Single Pipeline for Both Scan-Level and Map-Level Matching}\label{sec:outlier_robust_registration}

Before we present the details of our MSS, we briefly explain the outlier-robust 3D point cloud registration~\cite{yang2020graduated}, which is the core of our MSS.
We note that GNC-based global registration, which estimates the pose while rejecting outlier measurements simultaneously by gradually increasing the nonlinearity of the kernel~\cite{yang2020graduated},
can robustly and quickly align two point clouds while overcoming up to 70-80\% of outliers.
For this reason, in our previous work, we proposed an outlier-robust registration method specialized for urban environments, \textit{Quatro}~\cite{lim2022single, lim2023quatro++}, which is a variant of TEASER++~\cite{yang2020teaser}.
Quatro shows a high success rate even when distant source and target clouds with large pose discrepancies are provided.
These robust characteristics inspire the use of this methodology to estimate inter-session initial alignment.

In particular, we notice that Quatro is applicable not only to scan-level registration but also to map-level registration once correspondences from cloud points are given.
To establish correspondences between two maps or two submaps from each session, we use the fast point feature histogram~(FPFH)~\cite{rusu2009fpfh} because it can be easily used regardless of the scale of the point cloud.
By denoting the point cloud from the central session as $\mathcal{P}_C$ and that from the query session as $\mathcal{P}_{Q}$, the correspondence estimation $f(\cdot)$ can be formulated as follows:

\vseq
\begin{equation}
    \mathcal{A} = f\big(\mathcal{P}_C, \;\mathcal{P}_Q, \; \voxel\big),
    \label{eqn:matching}
\end{equation}

\noindent where $\voxel$ is the voxel size for voxel sampling of $\mathcal{P}_C$ and $\mathcal{P}_Q$ before the correspondence estimation; $\mathcal{A}$ is the set of correspondence pairs through the nearest neighbor search in the feature space of FPFH, which is followed by consistency-aware initial correspondence filtering to increase the inlier ratio~\cite{zhou2016fast}.
When extracting the FPFH feature descriptors for the voxel-sampled clouds, we set the radius for normal estimation to~$r_\text{normal}=3.5 \voxel$ and that for FPFH to $r_\text{FPFH}=5.0 \voxel$ to consider the voxel-sampled surrounding points.

\newcommand{\ith}{a}
\newcommand{\jth}{b}
Finally, we redefine $\mathcal{A}$ as $\mathcal{A}=\{(\ith, \jth)\}$, where $\ith$ and~$\jth$ are the indices of points in $\mathcal{P}_C$ and $\mathcal{P}_{Q}$, respectively, such that $\mathbf{p}_\ith \in \mathcal{P}_C$ and  $\mathbf{p}_\jth \in \mathcal{P}_Q$.
Then the objective function to estimate the relative rotation matrix~$\mathbf{R}\in\text{SO}(3)$ and translation vector~$\mathbf{t}\in\mathbb{R}^3$ is formulated as follows:

\begin{equation}
    \hat{\mathbf{R}}, \hat{\mathbf{t}} =\argmin_{\mathbf{R} \in \mathrm{SO}(3), \mathbf{t} \in \mathbb{R}^{3}}  \sum_{(\ith,\jth) \in \mathcal{A} \setminus \hat{\mathcal{O}}}\rho\Big( r(\mathbf{p}_\jth-\mathbf{R} \mathbf{p}_\ith-\mathbf{t})\Big),
    \label{eqn:final_goal}
\end{equation}

\noindent where $r(\cdot)$ is a squared residual function. Thus, \eqref{eqn:final_goal} means that the solver robustly estimates relative pose while rejecting the estimated outlier pairs $\hat{\mathcal{O}}$ simultaneously.
For brevity, we express the relative pose as $\hat{\mathbf{T}}\in \text{SE(3)}$ interchangeably with $\hat{\mathbf{R}}$ and $\hat{\mathbf{t}}$.

Note that \eqref{eqn:matching} and \eqref{eqn:final_goal} are used to estimate the initial alignment between the central and query sessions not only at the map-to-map level, but also at the scan-to-scan level, as detailed in Sections~\ref{sec:map_to_map} and~\ref{sec:scan_to_scan}.

\subsection{Outlier-Robust 3D Point Cloud Registration for Robust Initial Alignment Despite Large Pose Discrepancy}\label{sec:outlier_robust_registration_details}

Before we describe the pipeline of inter- and intra-session alignment, we briefly explain how outlier-robust registration operates.
In general, the pipeline for our outlier-robust registration mainly consists of three steps: a)~maximum clique inlier selection for initial outlier pruning, b)~GNC-based rotation estimation, and c)~component-wise translation estimation~(COTE).

First, maximum clique inlier selection~\cite{rossi2013pmc,pattabiraman2015pmc}, which is a graph-theoretic pruning approach~\cite{shi2021robin}, is applied to reject outlier correspondences in advance.
Finding the maximum clique among the correspondences is akin to identifying the subset that shows the largest consensus.
As a result, only those correspondences likely to be inliers remain after this process.

Next, the relative rotation and translation estimation is performed in a decoupled manner.
That is, following Horn's method~\cite{horn1987closed}, which estimates $\hat{\mathbf{R}}$ and $\hat{\mathbf{t}}$ independently, the relative rotation is calculated first in a translation-invariant space.
Subsequently, the translation is estimated in the next step.
Note that this decoupling-based approach relies on the assumption that the estimate of relative rotation is sufficiently accurate before proceeding to the translation estimation~\cite{yang2020teaser}.

\newcommand{\iprime}{{\ith^\prime}}
\newcommand{\jprime}{{\jth^\prime}}
Formally, by assuming that $\mathcal{A}$ is an ordered tuple, we define translation-invariant measurements~(TIMs) of $\mathcal{P}_Q$ and $\mathcal{P}_C$ as $\boldsymbol{\alpha}_k=\mathbf{p}_\ith - \mathbf{p}_{\iprime}$ and $\boldsymbol{\beta}_k = \mathbf{p}_\jth - \mathbf{p}_{\jprime}$,
where indices sets ($\ith$, $\jth$) and ($\iprime$, $\jprime$) correspond to the $n$-th and $(n+1)$-th elements of $\mathcal{A}$.
Note that this arrangement forms a chain. That is, when $n$ reaches the size of $\mathcal{A}$, the $(n+1)$-th element cycles back to the first element.

Because the relationship between $\mathbf{p}_\ith$ and $\mathbf{p}_\jth$ (as well as between $\mathbf{p}_{\iprime}$ and $\mathbf{p}_{\jprime}$) can be modeled as $\mathbf{p}_\jth =\mathbf{R} \mathbf{p}_\ith + \mathbf{t} + \boldsymbol{\epsilon}_{\ith, \jth}$, where $ \boldsymbol{\epsilon}_{\ith, \jth}$ indicates the noise term that follows a Gaussian distribution,
$\boldsymbol{\alpha}_{k}$ and $\boldsymbol{\beta}_{k}$ follow $\boldsymbol{\beta}_{k}=\mathbf{R}\boldsymbol{\alpha}_{k}+\boldsymbol{\epsilon}_{k}$ by canceling out the translation term~$\mathbf{t}$ if both $n$-th and $(n+1)$-th pairs are inliers.
Here, the $k$-th noise term in the translation-invariant space $\boldsymbol{\epsilon}_{k}$ is bounded by $-2\boldsymbol{\epsilon}_{\ith, \jth}~\leq~\boldsymbol{\epsilon}_{k}~\leq~2\boldsymbol{\epsilon}_{\ith, \jth}$~\cite{shi2021robin}.
Otherwise, $\boldsymbol{\beta}_{k} - \mathbf{R}\boldsymbol{\alpha}_{k}$ has a large, irregular error.

Based on these observations, the rotation is estimated in translation invariant space as follows:

\begin{equation}
    \hat{\mathbf{R}}=\underset{\mathbf{R} \in \mathrm{SO}(3)}{\argmin } \sum_{k=1}^{K} \min \Big( w_k r({\boldsymbol{\beta}}_{k}- \mathbf{R}{\boldsymbol{\alpha}}_{k}), \, \bar{c}^{2}\Big),
    \label{eqn:decouple_rot}
\end{equation}

\noindent where $K$ denotes the cardinality of correspondences in the translation-invariant space, $w_k$~is a weight term for the $k$-th residual term, $r(\cdot)$ is a squared residual function, and $\bar{c}$~is a user-defined parameter to truncate the effect of gross outliers.
Thus, (\ref{eqn:decouple_rot}) means that the undesirable effect is adjusted by the weight term $w_k$ when either $\boldsymbol{\alpha}_{k}$ or $\boldsymbol{\beta}_{k}$ is considered as an outlier,
and if the weighted squared residual term,~\ie $w_k r({\boldsymbol{\beta}}_{k}- \mathbf{R}{\boldsymbol{\alpha}}_{k})$, exceeds $\bar{c}^2$, the effect of the residual term is cut off.
By doing so, we can effectively suppress the undesirable effect of the potential outliers.
The detailed explanation of iteratively reweighted least squares with GNC, used to solve~\eqref{eqn:decouple_rot}, is described in Yang~\etalcite{yang2020teaser} and Lim~\etalcite{lim2023quatro++}.

Finally, component-wise translation estimation~(COTE)~\cite{yang2020teaser} is performed to estimate the relative translation.
As the name COTE suggests, the 3D relative translation is estimated element-wise as follows:

\begin{equation}
	^{l}{\hat{\mathbf{t}}}=\underset{^{l}\mathbf{t} \in \mathbb{R}}{\argmin } \sum_{(\ith,\jth) \in \mathcal{A}} \min \bigg( \frac{r({^{l}\mathbf{t}}-{^{l}\mathbf{v}_{\ith, \jth}})}{\sigma_{\ith, \jth}^2}, \, {\bar{c}^{2}} \bigg)
	\label{eqn:decouple_trans}
\end{equation}

\noindent where $\mathbf{v}_{\ith, \jth}=\mathbf{p}_{\jth}-{\hat{\mathbf{R}}} \mathbf{p}_{\ith}$ denotes the translation discrepancy, $\sigma_{\ith, \jth}$ is a user-defined noise bound, and $^l(\cdot)$ indicates the $l$-th element of a 3D vector.
That is, $l = 1,2,3$ and each value corresponds to $x$, $y$, and $z$ translation, respectively, in the ascending order.
Thus, \eqref{eqn:decouple_trans} operates similarly to consensus maximization~\cite{li2009consensus}, which involves performing a weighted summation based on the subset with the most overlap within the noise bounds.

In summary, our decoupling-based optimization decomposes (\ref{eqn:final_goal}) into (\ref{eqn:decouple_rot}) and (\ref{eqn:decouple_trans}).

\subsection{Map-to-Map Level Registration for Initial Alignment}\label{sec:map_to_map}

\newcommand{\reltf}{{\mathbf{T}}^{W_{C}}_{W_{Q}}}
\newcommand{\relesttf}{\hat{\mathbf{T}}^{W_{C}}_{W_{Q}}}
In \eqref{eqn:multi_session}, finding a valid $\mathcal{L}^\text{inter}$ becomes difficult for existing MSS approaches if $C$ is acquired by an Omni-LiDAR sensor, $Q$ is acquired by a Solid-LiDAR sensor, or vice versa, owing to the performance degradation of their LCD modules~\cite{kim2018scancontext}.
Consequently, this failure leads to a substantial inter-session misalignment~(see \secref{sec:sota_comparison}).
Thus, we address this initial alignment problem through map-to-map registration, which is tacitly considered impossible for direct registration with tens of millions of points in the presence of large pose discrepancies~\cite{bonanni20173d}.
However, we enable this using Quatro.

To perform map-to-map registration, we first accumulate map clouds for each session and then estimate the transformation matrix between the reference frames of $C$ and $Q$, $\relesttf$, where $W_{C}$ and $W_{Q}$ are the reference coordinate systems of the central and query sessions, respectively.
Formally, we define the accumulated map of the session~$s$, $\mathcal{M}_s$, as follows:

\begin{equation}
    \mathcal{M}_s=\bigcup_{i \in \langle N_s \rangle}{\Big\{\mathbf{T}_{s, i} \mathbf{p} \mid \mathbf{p} \in \mathcal{P}_{s,i} \Big\}},
    \label{eqn:accumulated_map}
\end{equation}

\noindent where $\mathbf{T}_{s, i}$ and $\mathcal{P}_{s, i}$ are the transformation matrix and point cloud corresponding to $\mathbf{x}_{s,i}$, respectively;
$\langle N_s \rangle=\{1, 2, \cdots, N_s\}$, where $N_s$ denotes the total frame number of the session~$s$.
Then, $\mathcal{M}_C$ and $\mathcal{M}_Q$ from~\eqref{eqn:accumulated_map} are taken as inputs of \eqref{eqn:matching} to estimate correspondences as $\mathcal{A}_m = f\big(\mathcal{M}_C, \, \mathcal{M}_Q, \, \vmap)$, where $\vmap$ denotes the voxel size for a map cloud.
Next, $\mathcal{A}_m$ is used to solve~\eqref{eqn:final_goal}; that is, $\mathcal{A} \leftarrow \mathcal{A}_m$, which finally outputs~$\relesttf$.

By denoting the transformation matrices for $\Delta_Q$ and $\Delta_C$ as $\mathbf{T}_{W_{Q}}$ and $\mathbf{T}_{W_{C}}$, respectively,
we can specify the initial transformation matrix for $\Delta_Q$ as $\mathbf{T}_{W_{Q}} = \mathbf{T}_{W_{C}} {\relesttf}$.
This contrasts with the conventional MSS methods that just set $\mathbf{T}_{W_{Q}} = \mathbf{T}_{W_{C}}$ because ${\relesttf}$ is unknown.

\begin{figure}[t!]
    \captionsetup{font=footnotesize}
    \centering
    \begin{subfigure}[b]{.43\textwidth}
        \centering
        \includegraphics[width=\textwidth]{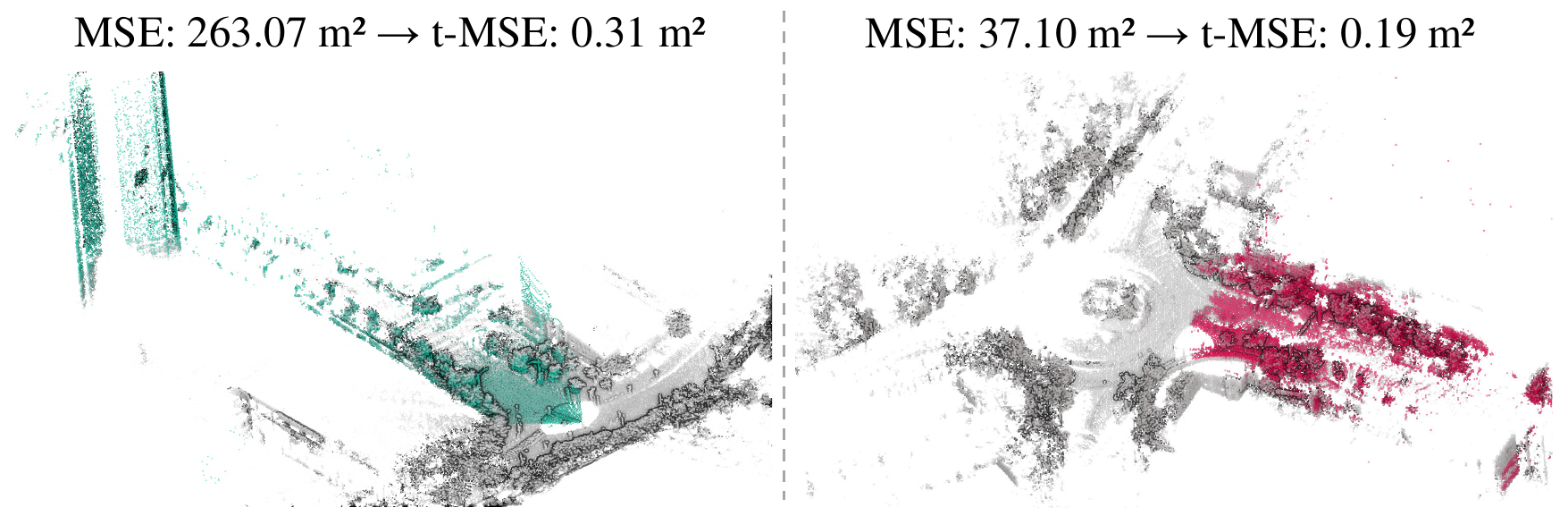}
    \end{subfigure}
    \caption{Original mean squared error~(MSE) and our truncated MSE~(t-MSE) of submap-to-submap registration results for inter-session loop constraints, where query~(from the query session) and target~(from the central session) point clouds are acquired by different type of LiDAR sensors.
        (L-R): The query and target are obtained by Livox Avia~(cyan) and Ouster OS2-128~(gray), and by Aeva Aeries\,\rom{2}~(magenta) and Ouster OS2-128~(gray), respectively~(best viewed in color).}
    \label{fig:trunc}
    \vsfig
\end{figure}

\subsection{Scan-to-Scan Level Inter-Session Loop Closing}\label{sec:scan_to_scan}

Once $C$ and $Q$ are initially aligned via $\relesttf$,
it is much easier to detect $\mathcal{L}^\text{inter}$ in \eqref{eqn:multi_session} because the poses of the scenes revisited across sessions are in close proximity to each other.
Therefore, for the $k$-th node of $Q$, we fetch the closest node in $C$, denoted as the $j$-th node, as a potential loop candidate.
Then, our inter-session loop detection is applied, which consists of four steps.

\newcommand{\dthres}{d_{\max}}
\newcommand{\dist}{d_n}
\newcommand{\dtrunc}{\text{t-MSE}}
\newcommand{\sbm}{\mathcal{S}}

First, to mitigate the impact of density differences between single scans acquired by heterogeneous LiDAR sensors,
we accumulate point clouds for $Q$ and $C$ by setting frame sets around the $k$-th and $j$-th nodes within range $T$, respectively.
Second, we estimate the relative pose between the two nodes in a coarse-to-fine manner using Quatro and local registration, such as the iterative closest point~(ICP)~\cite{lim2023quatro++}.
By denoting two submaps from the frame sets as $\sbm_{Q, k}$ and $\sbm_{C, j}$,
the relative pose is estimated by \eqref{eqn:final_goal} taking $f(\sbm_{C, j},\,\sbm_{Q, k}, \, \vsubmap)$ from~\eqref{eqn:matching} as correspondences, where $\vsubmap$ denotes the voxel size for a submap.
Then, ICP is applied as a fine alignment with maximum correspondence distance~$\dthres$.
That is, by letting the distance of the $n$-th pair between the warped query point in $\sbm_{Q, k}$ during the iteration and its closest point in $\sbm_{C, j}$ be $\dist$,
the correspondences whose $\dist$ is over $\dthres$ are not used in the optimization because these point-to-point correspondences are likely to be caused by the FoV differences of sensors, leading to undesirable pose errors.

Third, we introduce the \textit{truncated MSE~($\dtrunc$)} to additionally account for these differences in FoV, which is defined as follows:

\begin{equation}
    \dtrunc=\frac{1}{N_\text{inliers}} \sum_{n}\, [\dist \leq \dthres] \, \dist^2,
\end{equation}
\noindent where $\dist$ is recalculated by using the final estimated pose of our coarse-to-fine alignment, $N_\text{inliers}$ is the number of inliers that satisfies $\dist \leq \dthres$, and $[\cdot]$ is Iverson bracket, which outputs one if the condition is satisfied and zero otherwise.

Fourth, we reject the loop candidate whose $\dtrunc$ is larger than a threshold $\tau_\text{MSE}$, which means that two point clouds are not likely to be aligned.
This truncation may seem simple, but is extremely important when dealing with point clouds from heterogeneous LiDAR sensors.
Without such a truncation, the MSE between point clouds from heterogeneous LiDAR sensors could exhibit abnormally high values owing to FoV differences, making it impossible to determine whether the point clouds are well aligned or not, as shown in~\figref{fig:trunc}.

\begin{figure}[t!]
    \captionsetup{font=footnotesize}
    \centering
    \begin{subfigure}[b]{.43\textwidth}
        \centering
        \includegraphics[width=\textwidth]{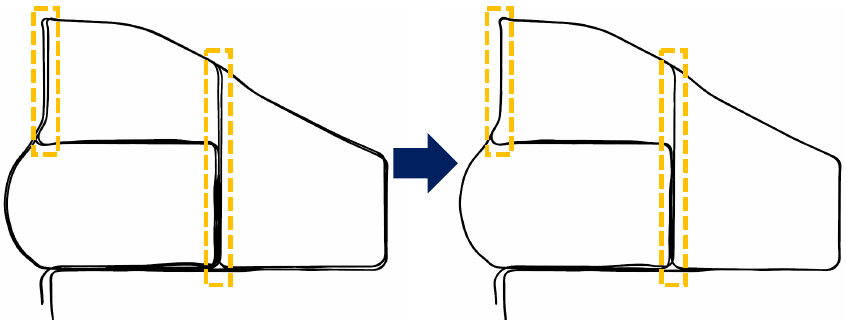}
    \end{subfigure}
    \caption{Before and after the application of our anchor node-based pose graph optimization to the trajectory in \texttt{DCC05} of the HeLiPR dataset~\cite{jung2023helipr}.
    Note that despite the erroneous trajectory from a single session, our approach successfully minimizes the trajectory errors, which are highlighted as orange dashed boxes~(best viewed in color).}
    \label{fig:b_and_a_optim}
    \vsfig
\end{figure}

\subsection{Global Map Building Across Multiple Sessions}\label{sec:global_map}

After completing the search for $\mathcal{L}^\text{inter}$, the objective function in \eqref{eqn:multi_session} is minimized, as presented in \figref{fig:overview}(d).
By doing so, poses are updated to build a consistent map across multiple sessions, resulting in more accurate poses, as shown in \figref{fig:b_and_a_optim}.
Consequently, by using updated poses of the nodes in the session~$s$,~$\hat{\mathbf{T}}_{s, i}$, and its updated anchor pose, $\hat{\mathbf{T}}_{W_s}$,
a global consistent map, $\mathcal{M}_\text{global}$, is constructed as follows:

\begin{equation}
    \mathcal{M}_\text{global}=\bigcup_{s \in \{C, Q\}} \Bigl\{  \bigcup_{i \in \langle N_s \rangle}{\big\{\hat{\mathbf{T}}_{W_s}\hat{\mathbf{T}}_{s, i} \mathbf{p} \mid \mathbf{p} \in \mathcal{P}_{s,i} \big\}} \Bigr\}.
    \label{eqn:final}
\end{equation}
 
\section{Experimental Evaluation}
\label{sec:exp}

The main focus of this study is to develop a simple and robust MSS approach applicable to heterogeneous LiDAR sensor setups.

We present the experiments that demonstrate the capabilities of the proposed method. The results of our experiments also support our key claims:
Our Multi-Mapcher (i)~successfully builds an accurate and consistent global map even though the sessions are acquired by heterogeneous LiDAR sensors,
(ii)~works robustly even in sessions with partial overlaps or dynamic changes between sessions,
and (iii)~offers faster and more efficient MSS performance than state-of-the-art scan-to-scan LCD-based MSS methods.

\subsection{Experimental Setup}

\begin{table}[t!]
    \centering
    \captionsetup{font=footnotesize}
    \caption{Parameters of our Multi-Mapcher.} {\scriptsize
    \setlength{\tabcolsep}{6pt}
        \begin{tabular}{lll}
            \toprule \midrule
            Parameter & Value & Description \\ \midrule
            $\vmap$    & 2.0$\,\mathrm{m}$ & Voxel sampling size at the map-to-map level \\
            $\vsubmap$ & 0.4$\,\mathrm{m}$ & Voxel sampling size at the submap-to-submap level \\
            $T$        & 20 & Window range of a submap for inter-session registration \\
            $\dthres$ & 2.0$\,\mathrm{m}$ & Maximum truncated correspondence distance in ICP \\
            $\tau_\text{MSE}$ & 0.4$\,\mathrm{m}^2$ & Threshold of our t-MSE-based filtering \\ \midrule
        \end{tabular}
    }
    \label{table:parameters}
    \vspace{-0.3cm}
\end{table}

\newcommand{\threepairs}{\texttt{O}-\texttt{L} & \texttt{O}-\texttt{A} & \texttt{L}-\texttt{A}}
\begingroup
\begin{table*}[t!]
    \captionsetup{font=footnotesize}
    \centering
    \caption{Quantitative comparison of MSS results using our inter-session alignment error~(iSAE), which is represented in the form of translation error~[m] / rotation error~[deg], in the HeLiPR dataset. The LiDAR sensor types used in the central $C$ and query $Q$ sessions are represented in the form of $C$-$Q$, which can be substituted with $\texttt{O}$ (Ouster OS2-128), $\texttt{L}$~(Livox Avia), and $\texttt{A}$~(Aeva Aeries~\rom{2}). The symbol \xmark$\,$ indicates failures of the MSS, where the inter-session translation and rotation differences exceed 100~m or 20 deg, respectively. \best{Green} and \second{blue} denote the best and second best performance,~respectively. $\texttt{S2Sub}$ and $\texttt{Sub2Sub}$ denote the scan-to-submap and submap-to-submap inter-session loop closing, respectively.}
    \setlength{\tabcolsep}{2pt}
    {\tiny
    \color{black}
        \begin{tabular}{l|ccccccccccccccc}
            \toprule \midrule
Sequence & \multicolumn{3}{c}{\texttt{DCC05}} & \multicolumn{3}{c}{\texttt{DCC06}} & \multicolumn{3}{c}{\texttt{KAIST05}} & \multicolumn{3}{c}{\texttt{KAIST06}} & \multicolumn{3}{c}{\texttt{Roundabout01}}  \\  \cmidrule(lr){2-4} \cmidrule(lr){5-7} \cmidrule(lr){8-10} \cmidrule(lr){11-13} \cmidrule(lr){14-16}
            LiDAR Info. & \threepairs & \threepairs & \threepairs & \threepairs & \threepairs \\ \midrule
            DiSCo-SLAM~\cite{huang2021disco} & \xmark & \xmark & \xmark & \xmark & \xmark & \xmark & \xmark & \xmark & \xmark & \xmark & \xmark & \xmark & \xmark & \xmark & 81.93\;/\;2.89 \\

            LT-mapper~\cite{kim2022lt} w/ SC~\cite{kim2018scancontext} & \xmark & \xmark & \xmark & \xmark & \xmark & \xmark & 12.80\;/\;3.71 & \xmark & 6.17\;/\;\second{2.31} & \xmark & \xmark & 5.40\;/\;0.49 & \xmark & 2.80\;/\;0.35 & 9.35\;/\;1.43 \\

            LT-mapper~\cite{kim2022lt} w/ STD~\cite{yuan2023std} & 11.61\;/\;1.23 & 2.10\;/\;0.61 & 7.29\;/\;\second{0.11} & 5.16\;/\;0.64 & \best{0.75}\;/\;\second{0.21} & \xmark & \xmark & 3.76\;/\;2.89 & \xmark & 4.82\;/\;0.56 & 3.12\;/\;\best{0.13} & 30.04\;/\;2.86 & 0.68\;/\;0.36 & 1.04\;/\;0.18 & \xmark \\

            Multi-Mapcher~(Ours) using \texttt{S2Sub} & \second{0.89}\;/\;\second{0.23} & \second{0.90}\;/\;\second{0.09} & \second{1.96}\;/\;0.57 & \second{1.56}\;/\;\best{0.08} & 2.34\;/\;2.25 & \second{2.01}\;/\;\best{0.21} & \second{1.16}\;/\;\best{0.22} & \second{2.94}\;/\;\second{2.83} & \second{3.58}\;/\;4.99  &  \second{0.83}\;/\;\best{0.10} & \second{2.65}\;/\;0.27 & \second{1.62}\;/\;\second{0.18} & \second{0.31}\;/\;\best{0.05} & \second{0.79}\;/\;\second{0.09} & \second{2.12}\;/\;\second{0.26} \\

            Multi-Mapcher~(Ours) using \texttt{Sub2Sub} & \best{0.83}\;/\;\best{0.08} & \best{0.81}\;/\;\best{0.08} & \best{0.49}\;/\;\best{0.07} & \best{1.30}\;/\;\second{0.25} & \second{1.13}\;/\;\best{0.19} & \best{1.35}\;/\;\second{0.38} & \best{1.11}\;/\;\second{0.30} & \best{0.51}\;/\;\best{0.12} & \best{1.98}\;/\;\best{0.23} & \best{0.81}\;/\;\second{0.15} & \best{0.97}\;/\;\second{0.26} & \best{1.12}\;/\;\best{0.04} & \best{0.23}\;/\;\second{0.14} & \best{0.57}\;/\;\best{0.06} & \best{1.61}\;/\;\best{0.08} \\ \midrule

            Sequence & \multicolumn{3}{c}{\texttt{Roundabout02}} & \multicolumn{3}{c}{\texttt{Roundabout03}} & \multicolumn{3}{c}{\texttt{Town01}} & \multicolumn{3}{c}{\texttt{Town02}} & \multicolumn{3}{c}{\texttt{Town03}} \\ \cmidrule(lr){2-4} \cmidrule(lr){5-7} \cmidrule(lr){8-10} \cmidrule(lr){11-13} \cmidrule(lr){14-16}
            LiDAR Info. & \threepairs & \threepairs & \threepairs & \threepairs & \threepairs \\ \midrule
            DiSCo-SLAM~\cite{huang2021disco} & \xmark & \xmark & 13.77\;/\;5.37 & \xmark & \xmark & 12.27\;/\;1.17 & \xmark & \xmark & 11.65\;/\;15.35 & \xmark & \xmark & \xmark & \xmark & \xmark & \xmark \\

            LT-mapper~\cite{kim2022lt} w/ SC~\cite{kim2018scancontext} & \xmark & 89.99\;/\;7.95 & 10.13\;/\;1.21 & \xmark & \xmark & \xmark & \xmark & \xmark & 2.25\;/\;0.53 & 7.03\;/\;2.14 & \xmark & \xmark & \xmark & \xmark & 4.57\;/\;0.38 \\

            LT-mapper~\cite{kim2022lt} w/ STD~\cite{yuan2023std} & 3.66\;/\;0.33 & \second{2.62}\;/\;\best{0.09} & 3.35\;/\;0.25 & 13.66\;/\;1.72 & 4.81\;/\;0.47 & 4.97\;/\;\second{0.47} & 6.30\;/\;2.78 & 5.77\;/\;1.26 & 19.28\;/\;1.69 & 11.49\;/\;6.40 & 20.10\;/\;2.26 & 6.95\;/\;1.21 & 25.49\;/\;2.37 & 2.19\;/\;0.59 & 12.74\;/\;4.29 \\

            Multi-Mapcher~(Ours) using \texttt{S2Sub} & \second{1.81}\;/\;\second{0.11} & 2.94\;/\;0.47 & \second{2.31}\;/\;\best{0.21} & \second{1.82}\;/\;\second{0.29} & \second{1.99}\;/\;\best{0.17} & \second{2.46}\;/\;0.80 & \second{1.65}\;/\;\best{0.04} & \second{2.78}\;/\;\best{0.10} & \second{3.19}\;/\;\best{0.38}  & \second{1.49}\;/\;\best{0.02} & \second{6.46}\;/\;\second{0.93} & \second{4.42}\;/\;\second{0.76} & \second{2.31}\;/\;\second{0.30} & \second{1.74}\;/\;\best{0.43} & \second{3.08}\;/\;\best{0.31} \\

            Multi-Mapcher~(Ours) using \texttt{Sub2Sub} & \best{1.02}\;/\;\best{0.05} & \best{1.77}\;/\;\second{0.38} & \best{1.93}\;/\;\second{0.23} & \best{0.89}\;/\;\best{0.07} & \best{0.79}\;/\;\second{0.42} & \best{0.72}\;/\;\best{0.34} & \best{0.91}\;/\;\second{0.07} & \best{2.13}\;/\;\second{0.19} & \best{1.66}\;/\;\second{0.47}  &  \best{0.57}\;/\;\second{0.06} & \best{5.76}\;/\;\best{0.30} & \best{2.86}\;/\;\best{0.48} & \best{0.79}\;/\;\best{0.03} & \best{1.57}\;/\;\second{0.48} & \best{2.14}\;/\;\second{0.51} \\
            \midrule\bottomrule
        \end{tabular}
    }
    \label{table:homo_session}
\end{table*}
\endgroup

In our experiments, our Multi-Mapcher was implemented based on iSAM2~\cite{kaess2012isam} of GTSAM~\cite{dellaert2018gtsam}.
To obtain the intra-session constraints, we used KISS-ICP-based, FAST-LIO2-based, and PV-LIO\footnote{PV-LIO is a reimplementation version of VoxelMap~\cite{yuan2022voxel} integrating IMU measurement. See {\texttt{https://github.com/HViktorTsoi/PV-LIO}}.}-based single session SLAM~\cite{vizzo2023kiss,wei2022fastlio,yuan2022voxel}, respectively.
By doing so, we demonstrate that our approach is agnostic to the choice of SLAM algorithms used for generating single-session trajectories and maps.
Note that the single session SLAM results also have inherent pose uncertainties, leading to inconsistent mapping. The parameter settings of the proposed method are listed in \tabref{table:parameters}.

To evaluate the performance of MSS approaches, we utilize four datasets: a)~HeLiPR~\cite{jung2023helipr}, b)~HILTI SLAM Challenge 2021~\cite{helm2022hilti}, c)~MulRan~\cite{kim2020mulran}, and d)~KITTI~\cite{geiger2012kitti} datasets.
Originally, the evaluation of the alignment between the query and central sessions is challenging because of the absence of ground truth between these sessions.
However, the HeLiPR dataset, acquired using four heterogeneous LiDAR sensors simultaneously, and HILTI SLAM Challenge 2021 (or HILTI 2021), which was also captured with multiple LiDAR sensors, enabled us to assess the performance of the MSS approaches.
This is because the relative poses between trajectories acquired using different LiDAR data within the same sequence are known.
Consequently, by treating data obtained by different LiDAR sensors in the same sequence as if acquired from different sessions and augmenting the poses of the query session with a large yaw rotation,
we were able to obtain a quantitative evaluation of the inter-session alignment.

For the quantitative evaluation of the alignment accuracy between sessions, we propose a novel metric called \textit{inter-session alignment error~(iSAE)},~$\mathbf{T}_\text{iSAE}$, which is defined as follows:

\newcommand{\mappingfigsize}{0.30}
\begin{figure*}[t!]
    \captionsetup{font=footnotesize, skip=5pt}
    \centering
    \begin{subfigure}[b]{\mappingfigsize\textwidth}
    	\captionsetup{skip=2pt}
        \includegraphics[width=1.0\textwidth]{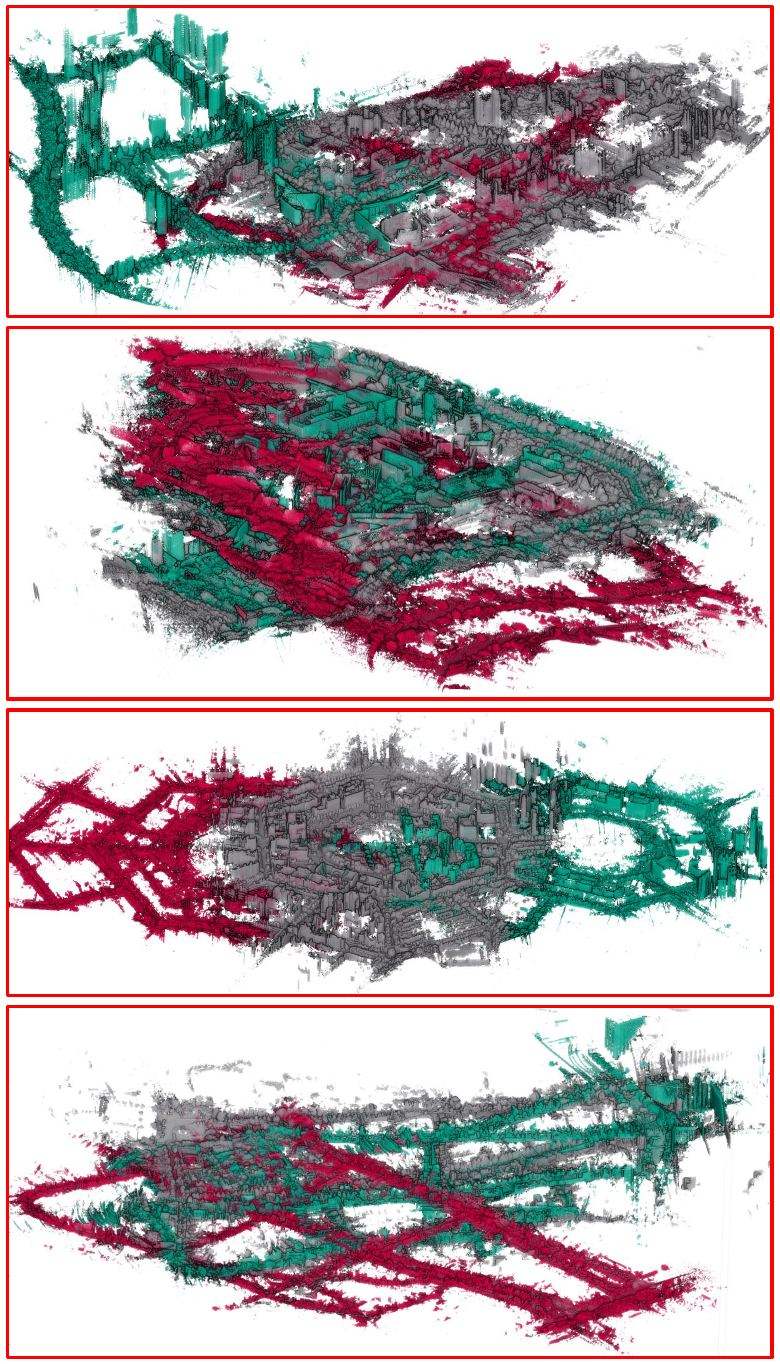}
        \caption{LT-mapper~\cite{kim2022lt} w/ SC~\cite{kim2018scancontext}}
    \end{subfigure}
    \begin{subfigure}[b]{\mappingfigsize\textwidth}
    	\captionsetup{skip=2pt}
        \includegraphics[width=1.0\textwidth]{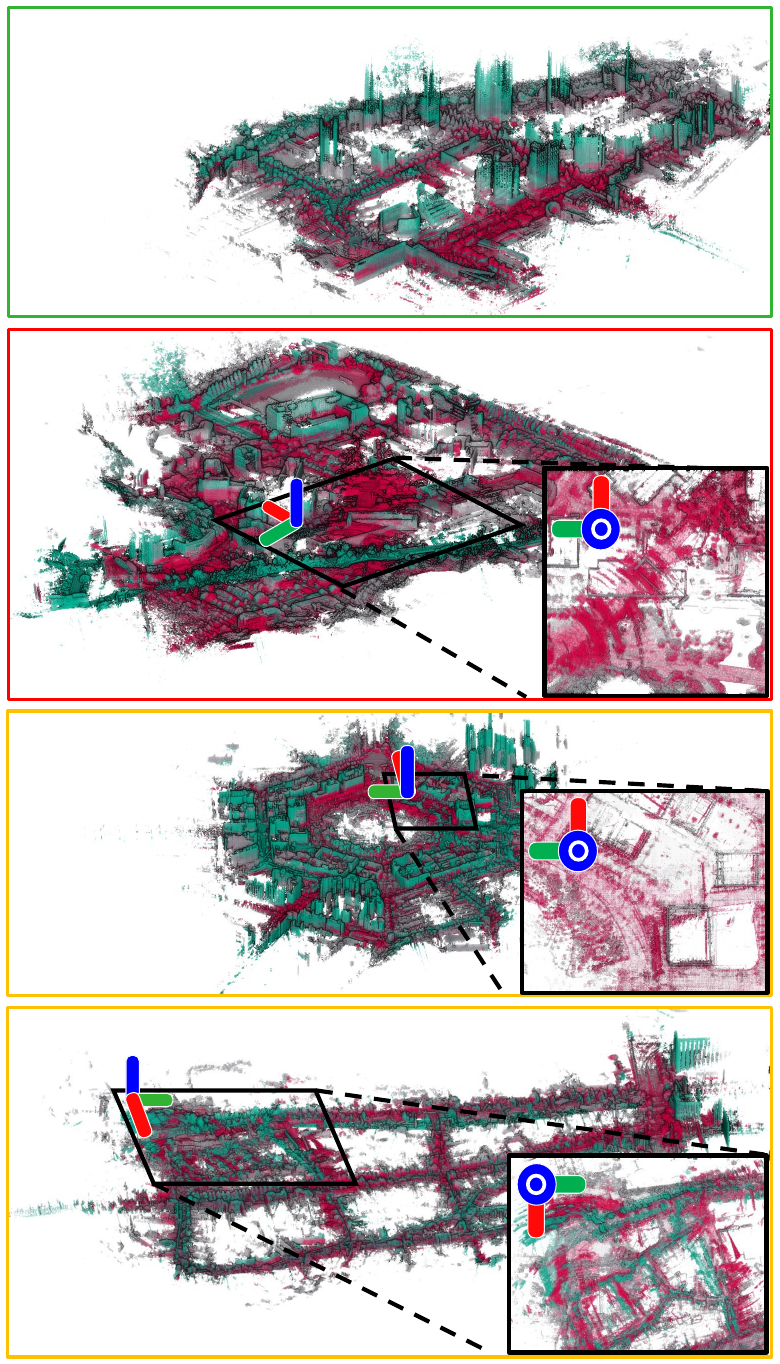}
        \caption{LT-mapper~\cite{kim2022lt} w/ STD~\cite{yuan2023std}}
    \end{subfigure}
    \begin{subfigure}[b]{\mappingfigsize\textwidth}
    	\captionsetup{skip=2pt}
        \includegraphics[width=1.0\textwidth]{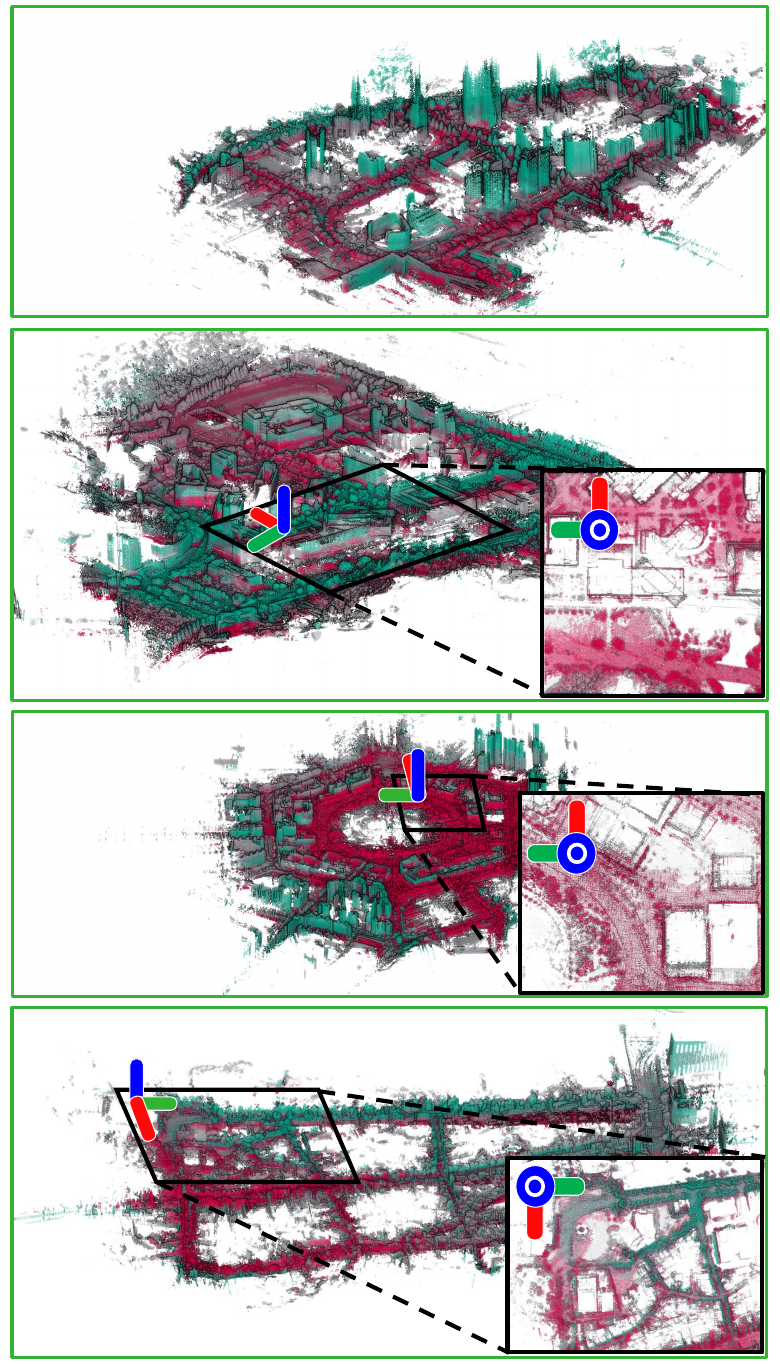}
        \caption{Multi-Mapcher (Ours)}
    \end{subfigure}
    \caption{(a)-(c) Qualitative comparison with the state-of-the-art MSS approaches on \texttt{DCC05}, \texttt{KAIST05}, \texttt{Roundabout03}, and \texttt{Town03} in the HeLiPR dataset (from top to bottom). Gray, dark cyan, and dark magenta colors indicate each session obtained by Ouster OS2-128~(\texttt{O}), Livox Avia~(\texttt{L}), and Aeva Aeries~\rom{2}~(\texttt{A}), respectively. Black boxes zoom in on specific areas to highlight the misalignment of LT-mapper~\cite{kim2022lt} with STD~\cite{yuan2023std} and to showcase the successful MSS results achieved by our Multi-Mapcher. Red, yellow, and green boxes, outlining subfigures, indicate the failure in both \texttt{O-A} and \texttt{O-L} sessions, failure in at least one of the sessions, and success in both sessions, respectively~(best viewed in color).}
    \label{fig:homo_pcd}
    \vsfig
\end{figure*} 
\newcommand{\loopfigsize}{0.32}
\begin{figure*}[t!]
    \captionsetup{font=footnotesize}
    \centering
    \begin{subfigure}[b]{\loopfigsize\textwidth}
        \centering
        \includegraphics[width=\textwidth]{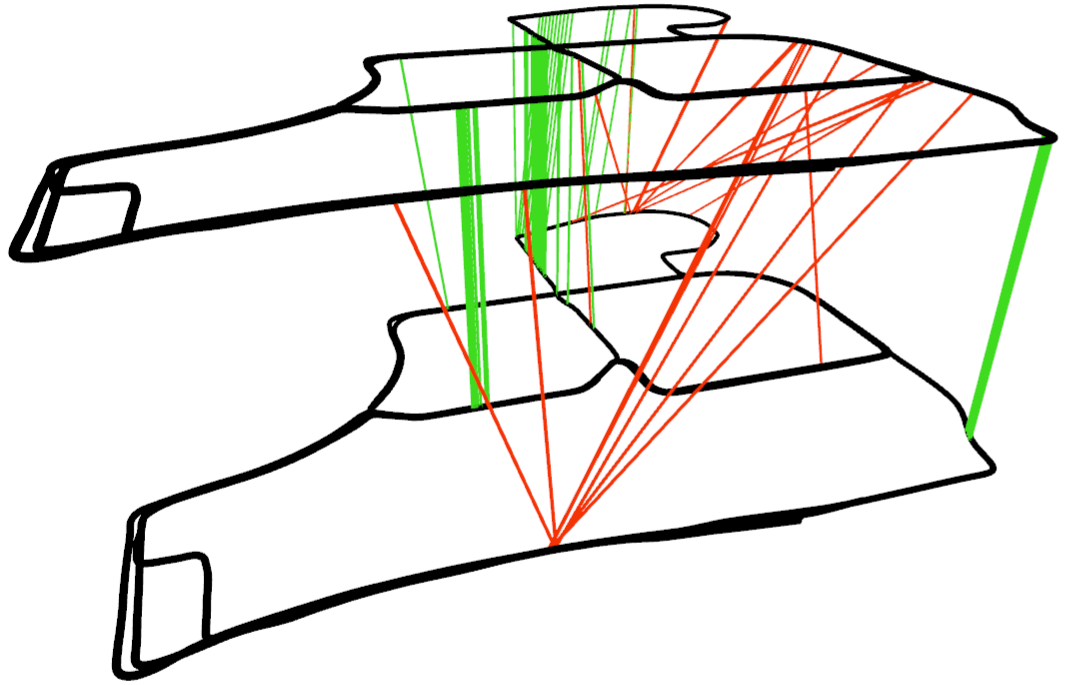}
    \end{subfigure}
    \hfill
    \begin{subfigure}[b]{\loopfigsize\textwidth}
        \centering
        \includegraphics[width=\textwidth]{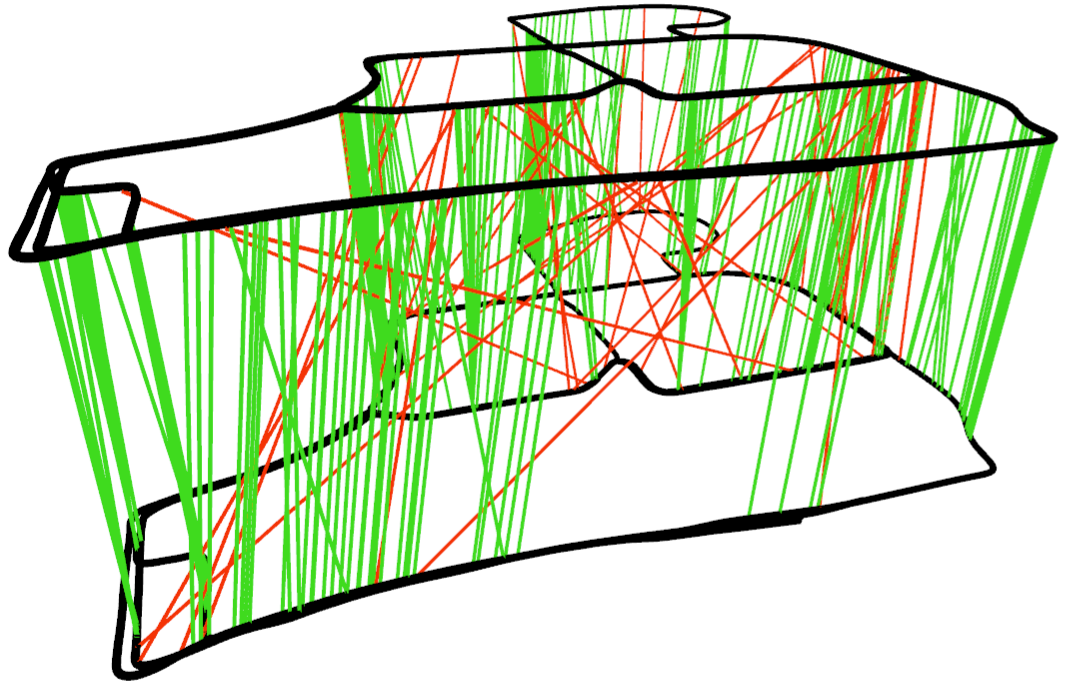}
    \end{subfigure}
    \hfill
    \begin{subfigure}[b]{\loopfigsize\textwidth}
        \centering
        \includegraphics[width=\textwidth]{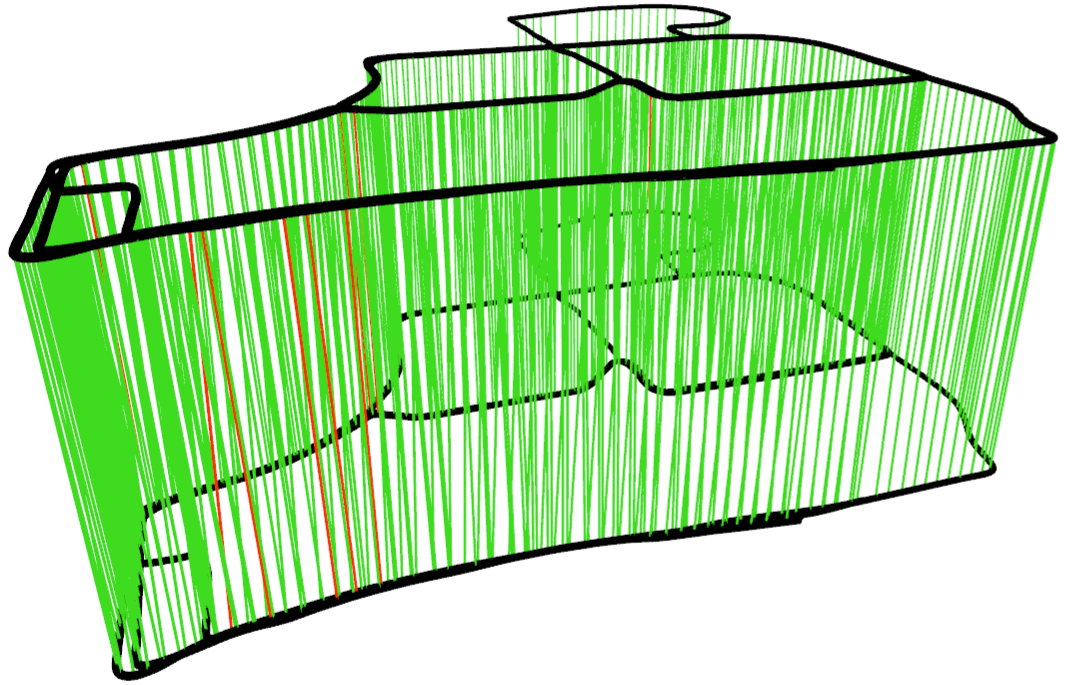}
    \end{subfigure}

    \vspace{0.5em} 

    \begin{subfigure}[b]{\loopfigsize\textwidth}
        \centering
        \includegraphics[width=\textwidth]{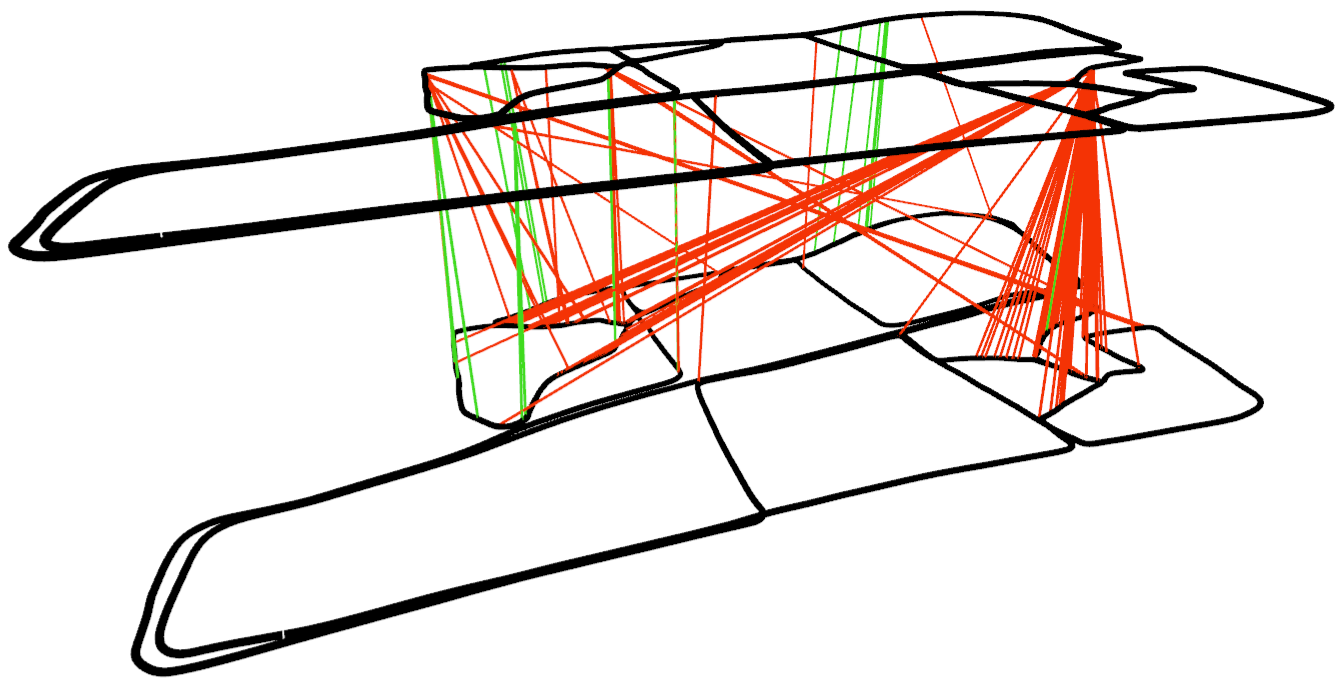}
        \caption{LT-mapper~\cite{kim2022lt} w/ SC~\cite{kim2018scancontext}}
    \end{subfigure}
    \hfill
    \begin{subfigure}[b]{\loopfigsize\textwidth}
        \centering
        \includegraphics[width=\textwidth]{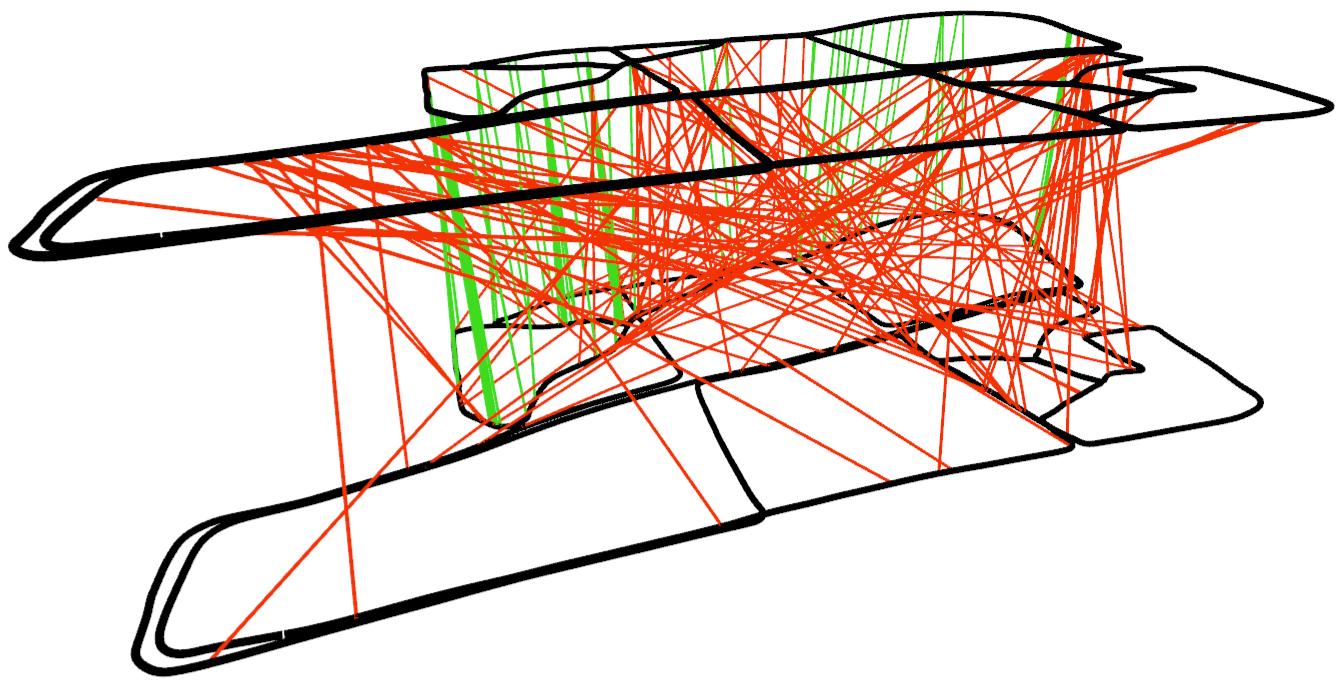}
        \caption{LT-mapper~\cite{kim2022lt} w/ STD~\cite{yuan2023std}}
    \end{subfigure}
    \hfill
    \begin{subfigure}[b]{\loopfigsize\textwidth}
        \centering
        \includegraphics[width=\textwidth]{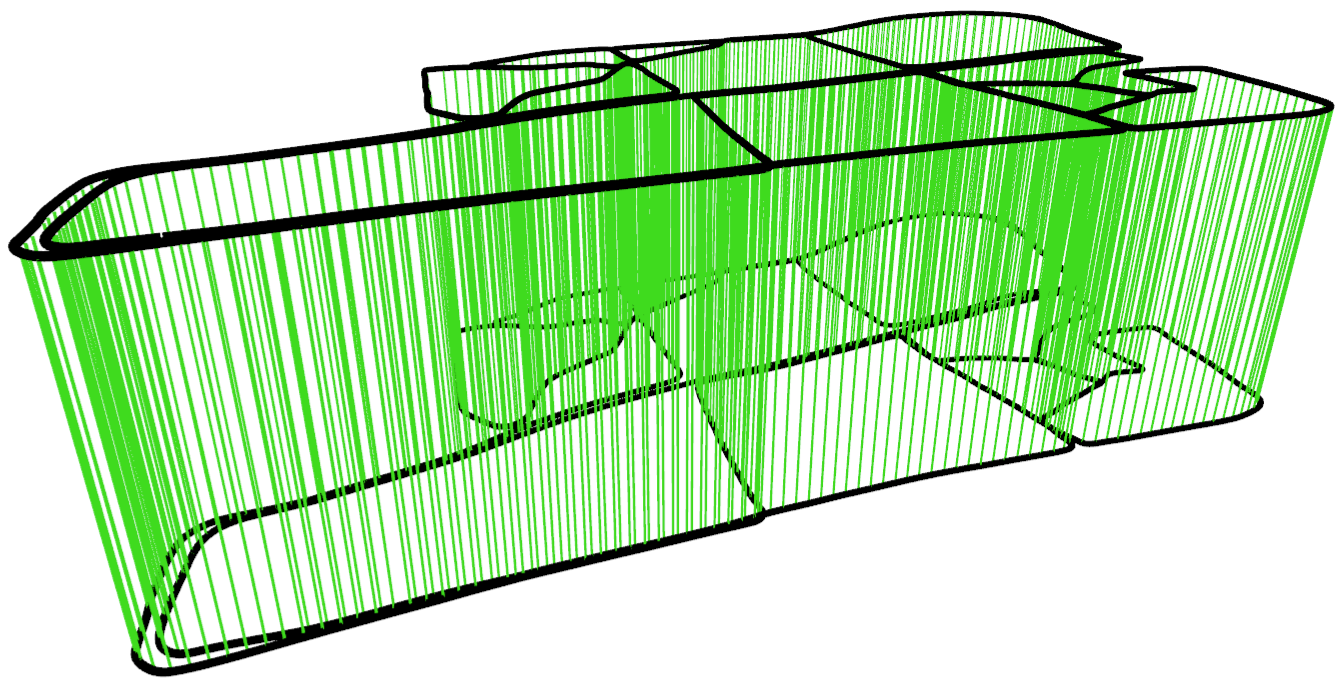}
        \caption{Multi-Mapcher (Ours)}
    \end{subfigure}
    \caption{Visualization of inter-session loop constraints of MSS results on \texttt{KAIST05}~(\texttt{O}-\texttt{L}) and \texttt{Town02}~(\texttt{O}-\texttt{L}) in the HeLiPR dataset (from top to bottom). The green and red lines indicate the true positive and false positive inter-session loop constraints, respectively. If both the translation and rotation errors of the loop constraints were less than 2~m and 10 deg, respectively, we considered them as true positive pairs, following the criteria in~\cite{lim2023quatro++} (best viewed in color).}
    \label{fig:loop_viz}
\end{figure*}

\newcommand{\gtsubscript}{\text{GT}}
\newcommand{\estsubscript}{\text{EST}}
\begin{equation}
  \mathbf{T}_\text{iSAE} =  (\hat{\mathbf{T}}^{W_{C, \gtsubscript}}_{W_{C, \estsubscript}})^{-1} \mathbf{T}^{W_{C, \gtsubscript}}_{W_{Q, \gtsubscript}} \hat{\mathbf{T}}^{W_{Q, \gtsubscript}}_{W_{Q, \estsubscript}} = \hat{\mathbf{T}}^{W_{C, \estsubscript}}_{W_{Q, \estsubscript}},
\end{equation}

\noindent where $\hat{\mathbf{T}}^{W_{C, \gtsubscript}}_{W_{C, \estsubscript}}$ and $\hat{\mathbf{T}}^{W_{Q, \gtsubscript}}_{W_{Q, \estsubscript}}$ are the estimated alignment transformation matrices between the ground truth and estimated poses of $C$ and $Q$ by Kabsch-Umeyama algorithm~\cite{umeyama1991least}, respectively,
and $\mathbf{T}^{W_{C, \gtsubscript}}_{W_{Q, \gtsubscript}}$ is known transformation matrix between actual world frames $C$ and $Q$.
In addition, we used the absolute pose error~(APE) to measure errors within the trajectory of each session using the EVO trajectory evaluation tool~\cite{grupp2017evo}.

Likewise, the HILTI 2021 dataset, acquired with the Ouster OS0-64 and Livox MID-70, was utilized to demonstrate the performance of MSS in various environments.
We also qualitatively analyze the impact of low- and high-dynamic changes between two sessions with significant time differences in data acquisition to demonstrate the applicability of long-term MSS.
To this end, we perform MSS between the \texttt{KAIST05} sequence of the HeLiPR dataset and the \texttt{KAIST01} sequence of the MulRan dataset.
This is because \texttt{KAIST01} was acquired four years before \texttt{KAIST05} sequence was collected.
Finally, the KITTI dataset acquired using the Velodyne HDL-64E was employed to demonstrate the robustness of our method in more partially overlapped scenarios.

\subsection{Heterogeneous LiDAR MSS Performance Comparison With State-of-the-Art Approaches}\label{sec:sota_comparison}

First, our Multi-Mapcher was quantitatively and qualitatively compared with state-of-the-art methods, namely, DiSCo-SLAM~\cite{huang2021disco}, LT-mapper~\cite{kim2022lt} with Scan Context~(w/ SC)~\cite{kim2018scancontext}, and LT-mapper with stable triangle descriptor~(w/ STD)~\cite{yuan2023std}.
Note that we fine-tuned the parameters of state-of-the-art approaches. Otherwise, these approaches fail to find inter-session loop candidates, resulting in catastrophic failure of the inter-session initial alignment.

\begin{figure*}[t!]
	\captionsetup{font=footnotesize}
	\centering
	\begin{subfigure}[b]{\textwidth}
		\centering
		\includegraphics[width=\textwidth]{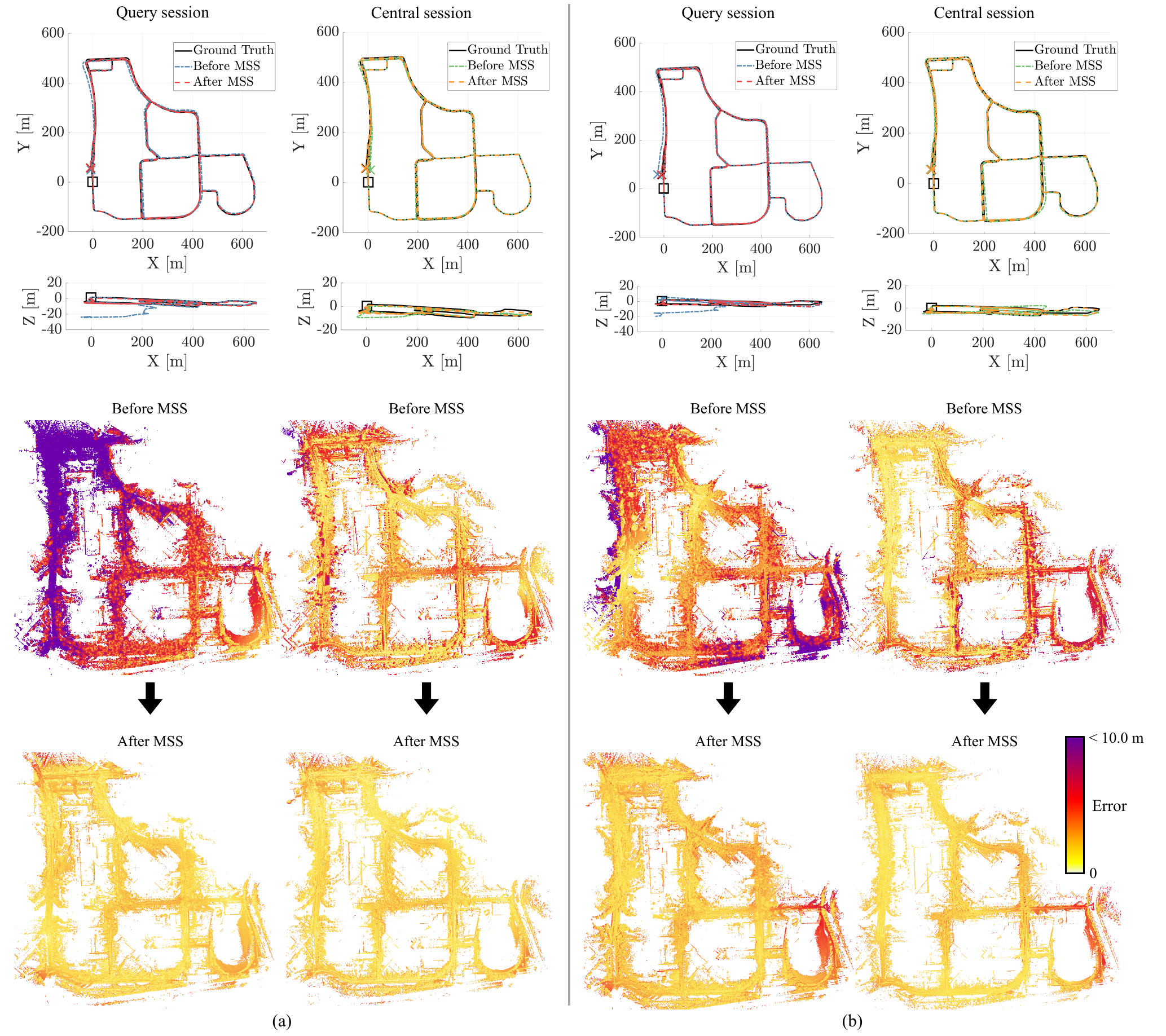}
	\end{subfigure}
	\caption{(a)-(b) Qualitative analysis of Multi-Mapcher results on single session SLAM outputs based on KISS-ICP~\cite{vizzo2023kiss} and FAST-LIO2~\cite{wei2022fastlio} using \texttt{KAIST05}~(\texttt{L-A}) of the HeLiPR dataset.
    For each subfigure, the results of the query session (left) and central session (right) are shown.
    From top to bottom, a top view and a side view of trajectories are presented, and visualization of the point-to-point error differences between the estimated accumulated map and the ground truth point cloud map before and after the application of MSS is presented.
    Note that Multi-Mapcher significantly improves the alignment and thus reduces the errors between sessions~(best viewed in color).}
	\label{fig:diff_mss}
	\vsfig
\end{figure*}

As shown in \tabref{table:homo_session} and \figref{fig:homo_pcd}, we demonstrate that our Multi-Mapcher accurately aligns heterogeneous LiDAR sessions.
While the state-of-the-art LCD-based approaches occasionally showed promising performance when inter-session loop candidates were successfully detected,
these methods failed in inter-session alignment owing to too many false positive inter-session loop candidates in the heterogeneous LiDAR sensor setups, as presented in \figref{fig:loop_viz}.
By contrast, our Multi-Mapcher showed substantially better performance in heterogeneous LiDAR sensor setups by reducing the dependency on an LCD module, achieving successful MSS across all sequences~(\tabref{table:homo_session}) with smaller APEs~(\tabref{table:pose_error}).
Consequently, as shown in ~\figref{fig:diff_mss}, our Multi-Mapcher successfully achieved consistent trajectories and accurate global maps.

\begin{figure*}[t!]
	\captionsetup{font=footnotesize}
	\centering
	\begin{subfigure}[b]{.30\textwidth}
		\centering
		\includegraphics[width=\textwidth]{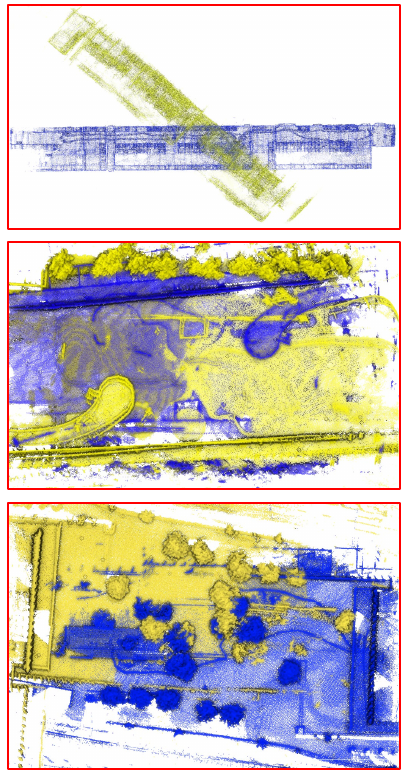}
		\caption{LT-mapper~\cite{kim2022lt} w/ SC~\cite{kim2018scancontext}}
		\label{fig:hilti_sc}
	\end{subfigure}
	\begin{subfigure}[b]{.30\textwidth}
		\centering
		\includegraphics[width=\textwidth]{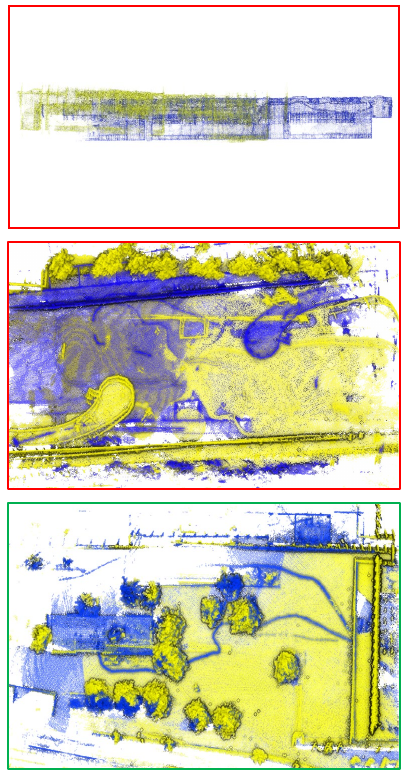}
		\caption{LT-mapper~\cite{kim2022lt} w/ STD~\cite{yuan2023std}}
		\label{fig:hilti_std}
	\end{subfigure}
	\begin{subfigure}[b]{.30\textwidth}
		\centering
		\includegraphics[width=\textwidth]{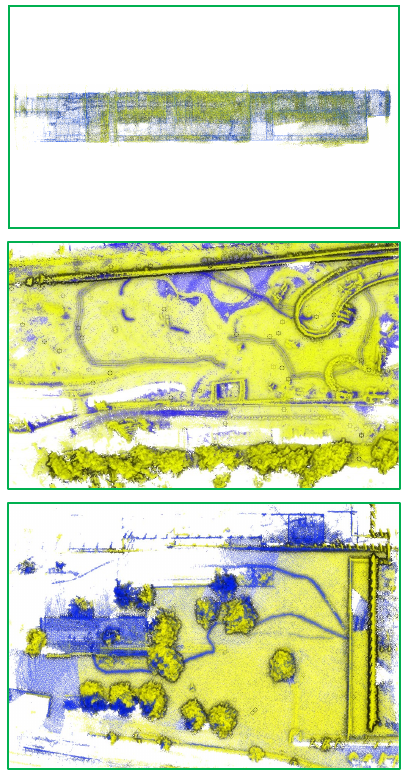}
		\caption{Multi-Mapcher (Ours)}
		\label{fig:hilti_ours}
	\end{subfigure}
	\caption{(a)-(c) Qualitative comparison with the state-of-the-art MSS approaches on \texttt{Basement4}, \texttt{Construction Site Outdoor2}, and \texttt{Campus2} of the HILTI SLAM Challenge 2021 dataset~(from top to bottom). Blue and yellow points represent the central~(captured by the Ouster OS0-64) and query~(captured by the Livox MID-70) session clouds, respectively. Red and green boxes, outlining subfigures, indicate the failure or success of the MSS, where the inter-session translation and rotation differences exceed 10~m or 20~deg, respectively~(best viewed in color).}
	\label{fig:hilti_fig}
\end{figure*}

In addition, as shown in~\tabref{table:hilti_ape} and~\figref{fig:hilti_fig}, we evaluated our Multi-Mapcher in both indoor and outdoor environments with heterogeneous LiDAR sensors using the HILTI 2021 dataset.
The state-of-the-art LCD-based approaches failed in some sequences owing to the absence of any valid descriptor matches or false positive inter-session loop candidates in the heterogeneous LiDAR sensor setups~(Figs.~\ref{fig:hilti_fig}(a) and ~\ref{fig:hilti_fig}(b)).
In contrast, our Multi-Mapcher successfully aligned heterogeneous LiDAR sessions across diverse environments~(\figref{fig:hilti_fig}(c)) and achieved lower APEs~(\tabref{table:hilti_ape}).
These results support that our approach is not only robust in heterogeneous LiDAR sensor setups but also highly applicable in various scenarios.

Furthermore, as shown in \tabref{table:homo_session}, submap-to-submap inter-session loop closing~(referred to as \texttt{Sub2Sub}) mostly outperformed scan-to-submap-based loop closing~(referred to as \texttt{S2Sub}).
This supports that our submap-to-submap-based inter-session loop closing effectively mitigates the geometrical discrepancies present in single scans from heterogeneous LiDAR sensors.
Therefore, we conclude that our Multi-Mapcher is a more robust LiDAR sensor-agnostic approach than the existing LCD-based MSS methods.

\subsection{Robustness in Partially Overlapped Sessions}

\begin{figure}[t!]
    \captionsetup{font=footnotesize}
    \centering
\begin{subfigure}[b]{.15\textwidth}
        \centering
        \includegraphics[width=\textwidth]{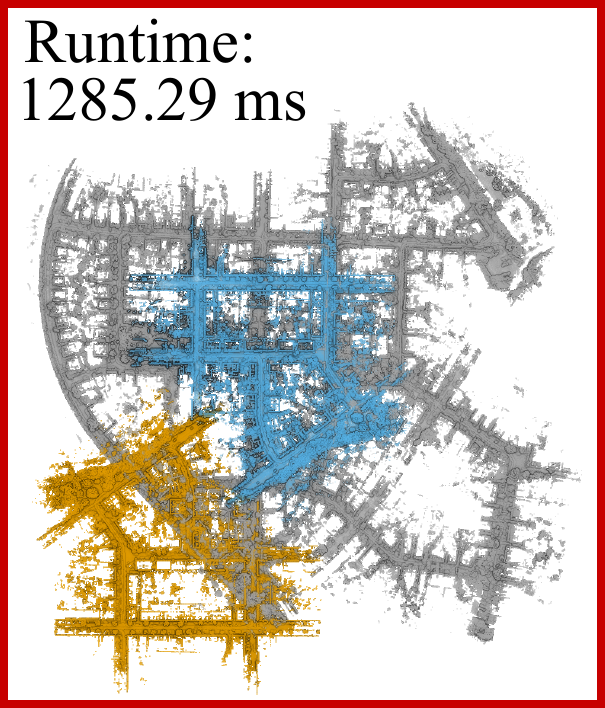}
        \caption{SAC-IA~\cite{rusu2009fpfh}}
        \label{fig:reg_sacia}
    \end{subfigure}
    \begin{subfigure}[b]{.15\textwidth}
        \centering
        \includegraphics[width=\textwidth]{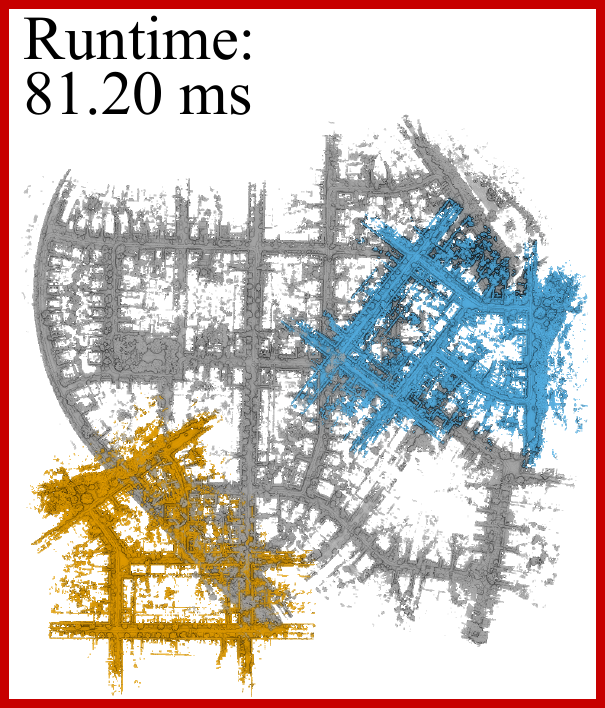}
        \caption{PCR-99~\cite{lee2024pcr}}
        \label{fig:reg_pcr99}
    \end{subfigure}
    \begin{subfigure}[b]{.15\textwidth}
        \centering
        \includegraphics[width=\textwidth]{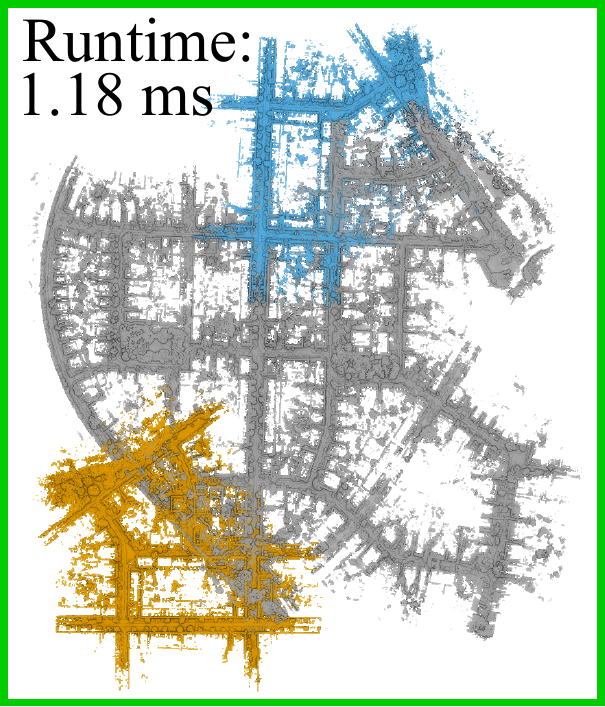}
        \caption{Multi-Mapcher (Ours)}
        \label{fig:reg_quatro}
    \end{subfigure}
    \caption{(a)-(c)~Comparison of the initial alignment results (cyan) with (a)~SAC-IA~\cite{rusu2009fpfh} and (b)~PCR-99~\cite{lee2024pcr}, which are robust registration methods, and (b) our Multi-Mapcher in partially overlapped cases using the KITTI dataset: Seq.~\texttt{00}~(orange) and Seq.~\texttt{07}~(gray) are used as the query and central sessions. Note that Multi-Mapcher is capable of robust registration even in the partially overlapped sessions.
    Red and green boxes, outlining subfigures, indicate the failure and success in registration, respectively. Note that there is only 14\% of overlap in Seq.~\texttt{00} of the KITTI dataset~(best viewed in color).}
    \label{fig:registration}
\end{figure}

Next, as shown in \figref{fig:registration}, we demonstrate the robustness of our approach in MSS with only partially overlapped sessions as inputs.
Note that other robust registration approaches, SAC-IA~\cite{rusu2009fpfh}~(\figref{fig:registration}(a)) and PCR-99~\cite{lee2024pcr}~(\figref{fig:registration}(b)), failed to align the partially overlapped sessions, whereas Quatro in Multi-Mapcher successfully aligned them~(Fig.~\ref{fig:registration}(c)).
Therefore, this result supports our second claim that our approach is even robust to partially overlapped scenarios.

\begin{table}[t!]
       \centering
       \captionsetup{font=footnotesize}
       \setlength{\tabcolsep}{6.0pt}
       \caption{Absolute pose errors~(APEs) of MSS results on \texttt{KAIST06}, \texttt{Roundabout02}~(\texttt{RA02}) and \texttt{Town03} of the HeLiPR dataset. The columns of $C$ and $Q$ represent the RMSE for the central and query sessions, respectively.
       The + symbol represents the results after the application of each multi-session SLAM approach. For a fair comparison, sequences where all three algorithms succeeded, i.e. none of them marked with \xmark\, in \tabref{table:homo_session}, are evaluated using APEs.}
       {\tiny
               \begin{tabular}{l|cccccc}
                       \toprule \midrule
                       \multirow{2}{*}[-0.3em]{Method} & \multicolumn{2}{c}{\texttt{KAIST06}~(\texttt{L}-\texttt{A})} &
						\multicolumn{2}{c}{\texttt{RA02}~(\texttt{L}-\texttt{A})} & \multicolumn{2}{c}{\texttt{Town03}~(\texttt{L}-\texttt{A})} \\ \cmidrule(lr){2-3} \cmidrule(lr){4-5} \cmidrule(lr){6-7}
						& $C$ & $Q$ & $C$ & $Q$ & $C$ & $Q$ \\ \midrule
						Single session SLAM with KISS-ICP~\cite{vizzo2023kiss} & 16.54 & 17.50 & 19.33 & 35.09 & 37.48 & 30.97 \\
						+ LT-mapper~\cite{kim2022lt} w/ SC~\cite{kim2018scancontext} & 12.98 & 14.78 & 23.98 & 24.71 & 51.24 & 43.16 \\
						+ LT-mapper~\cite{kim2022lt} w/ STD~\cite{yuan2023std} & 16.50 & 17.31 & 14.15 & 18.13 & 41.08 & 33.31 \\
						+ Multi-Mapcher (Ours) & \textbf{3.09} & \textbf{4.08} & \textbf{8.31} & \textbf{8.92} & \textbf{17.03} & \textbf{18.81}  \\ \midrule

                       Single session SLAM with FAST-LIO2~\cite{wei2022fastlio} & 5.79 & 16.52 & 4.82 & 34.79 & 9.24 & 37.26 \\
                       + LT-mapper~\cite{kim2022lt} w/ SC~\cite{kim2018scancontext} & 10.59 & 5.14 & 9.47 & 20.11 & 41.39 & 44.05 \\
                       + LT-mapper~\cite{kim2022lt} w/ STD~\cite{yuan2023std} & 40.33 & 15.14 & 16.40 & 15.87 & 58.09 & 59.02 \\
                       + Multi-Mapcher (Ours) & \textbf{2.96} & \textbf{4.06} & \textbf{4.45} & \textbf{12.17} & \textbf{3.32} & \textbf{8.20}  \\
                       \midrule \bottomrule
               \end{tabular}
       }
       \label{table:pose_error}
\end{table}

\begin{table}[t!]
	\centering
	\captionsetup{font=footnotesize}
	\setlength{\tabcolsep}{6.0pt}
	\caption{Absolute pose errors~(APEs) of MSS results on \texttt{Basement4}, \texttt{Construction Site Outdoor2}, and \texttt{Campus2} of the HILTI SLAM Challenge 2021 dataset. Here, the symbols \texttt{O} and \texttt{M} indicate that we used the sessions acquired by Ouster OS0-64 and Livox MID-70 LiDAR sensors, respectively. The columns of $C$ and $Q$ represent the RMSE for the central and query sessions, respectively.
		The + symbol represents the results after the application of each multi-session SLAM approach. The symbol \xmark~denotes failures of the MSS, where the inter-session translation and rotation differences exceed 10~m or 20~deg, respectively.}
	{\tiny
		\begin{tabular}{l|cccccc}
			\toprule \midrule
			\multirow{2}{*}[-0.3em]{Method} & \multicolumn{2}{c}{\texttt{Basement4}~(\texttt{O}-\texttt{M})} &
			\multicolumn{2}{c}{\texttt{CSO2}~(\texttt{O}-\texttt{M})} & \multicolumn{2}{c}{\texttt{Campus2}~(\texttt{O}-\texttt{M})} \\ \cmidrule(lr){2-3} \cmidrule(lr){4-5} \cmidrule(lr){6-7}
			& $C$ & $Q$ & $C$ & $Q$ & $C$ & $Q$ \\ \midrule

			Single session SLAM with PV-LIO & 0.25 & 0.38 & 0.21 & \textbf{0.12} & 0.13 & \textbf{0.15} \\
			+ LT-mapper~\cite{kim2022lt} w/ SC~\cite{kim2018scancontext} & \xmark & \xmark & \xmark & \xmark & \xmark & \xmark \\
			+ LT-mapper~\cite{kim2022lt} w/ STD~\cite{yuan2023std} & \xmark & \xmark & \xmark & \xmark & 0.14 & 0.16 \\
			+ Multi-Mapcher (Ours) & \textbf{0.18} & \textbf{0.20} & \textbf{0.15} & 0.13 & \textbf{0.08} & \textbf{0.15}  \\
			\midrule \bottomrule
		\end{tabular}
	}
	\label{table:hilti_ape}
\end{table}

\subsection{Robustness Against Low- and High-Dynamic Changes}

\begin{figure}[t!]
	\captionsetup{font=footnotesize}
	\centering
	\begin{subfigure}[b]{0.48\textwidth}
		\centering
		\includegraphics[width=\textwidth]{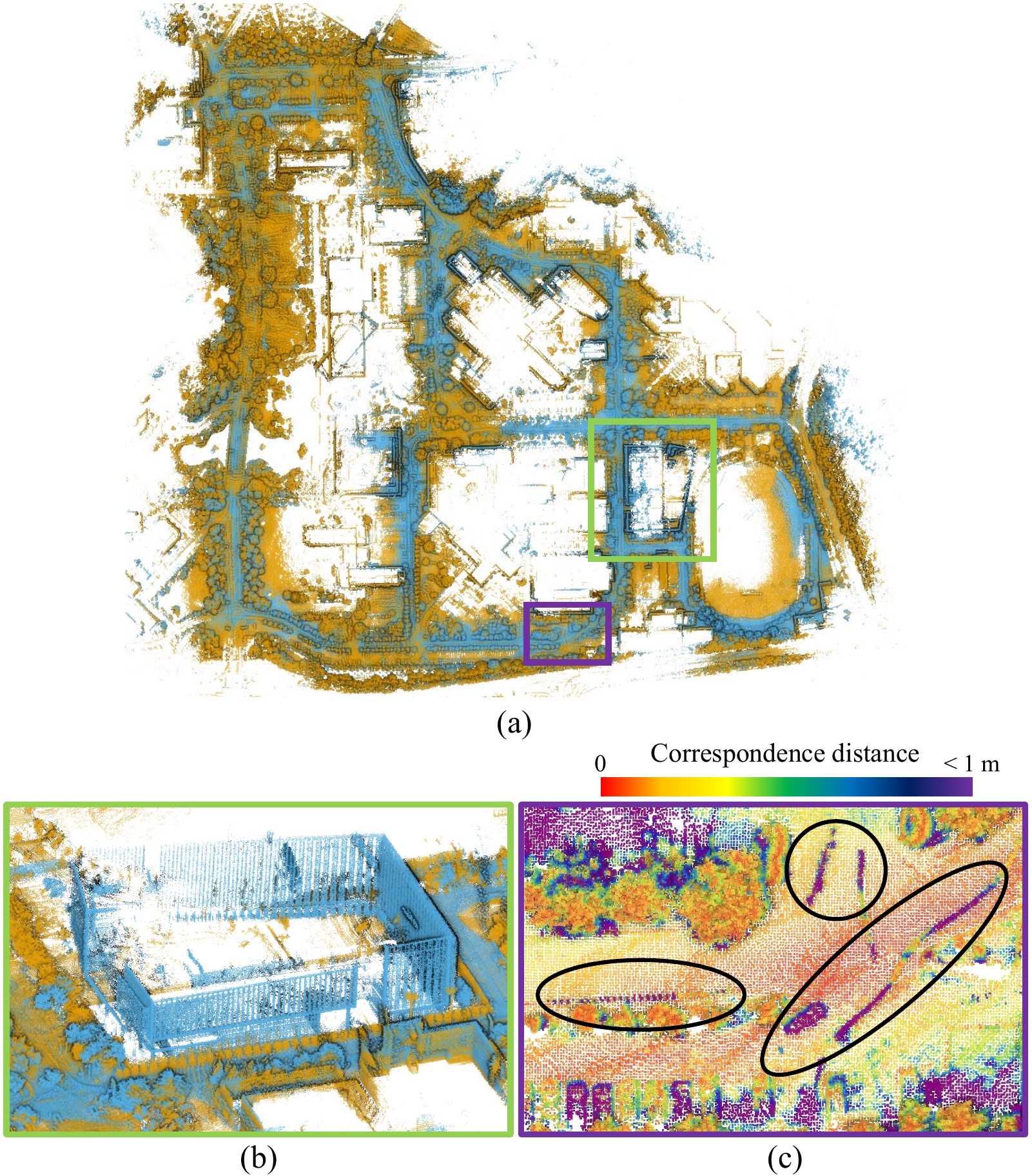}
	\end{subfigure}
	\caption{Multi-session SLAM results of Multi-Mapcher in low- and high-dynamic environments. \texttt{KAIST01} of MulRan dataset captured by an Ouster OS1-64 LiDAR sensor, was used as the query session~(orange) and \texttt{KAIST05} of HeLiPR dataset captured by LiVOX Avia~(\texttt{L}) was used as the central session~(cyan). (a)~The result of the merged map from two sessions, with colored boxes marking regions of (b)~low-dynamic~(\ie~new building construction between two sessions) and (c)~high-dynamic changes~(\ie~points from moving objects in each session). The solid black circles in (c) highlight the existence of high-dynamic changes, and rainbow colors indicate the closest distances between points in the query and central session clouds for clarity (best viewed in color).}
	\label{fig:mos_fig}
	\vsfig
\end{figure}

One notable aspect of our Multi-Mapcher is sufficient robustness against both low- and high-dynamic changes.
As shown in \figref{fig:mos_fig}, our method robustly performed MSS even in cases where a new building was constructed between sessions and when moving objects were observed in each session.
These results imply that our approach can achieve inter-session and intra-session alignment even in the presence of occupancy discrepancies that might lead to erroneous feature matching.

This robustness is attributed to the characteristics of the outlier-robust registration, as explained in \secref{sec:outlier_robust_registration}.
Our registration pipeline can tolerate these noisy correspondences with up to 99\% of outliers,
making it resilient enough to handle undesirable outliers within the putative correspondences in these low- and high-dynamic scenarios.
Thus, even though a new building was constructed~(\figref{fig:mos_fig}(b)), consisting of tens of thousands of cloud points and potentially triggering a large number of outliers,
our approach achieved precise inter- and intra-session alignment.
Similarly, even though the 3D cloud points from moving objects were observed~(\figref{fig:mos_fig}(c)),
their impact on registration performance was negligible because the ratio of outliers from these moving objects was not as significant as that from low-dynamic changes.

Therefore, this capability to mitigate the impact of low- and high-dynamic objects highlights the robustness of our Multi-Mapcher in handling scene changes of real-world environments.

\subsection{Ablation Study}

\begin{figure}[t!]
	\captionsetup{font=footnotesize}
	\centering
	\begin{subfigure}[b]{.24\textwidth}
		\centering
		\includegraphics[width=\textwidth]{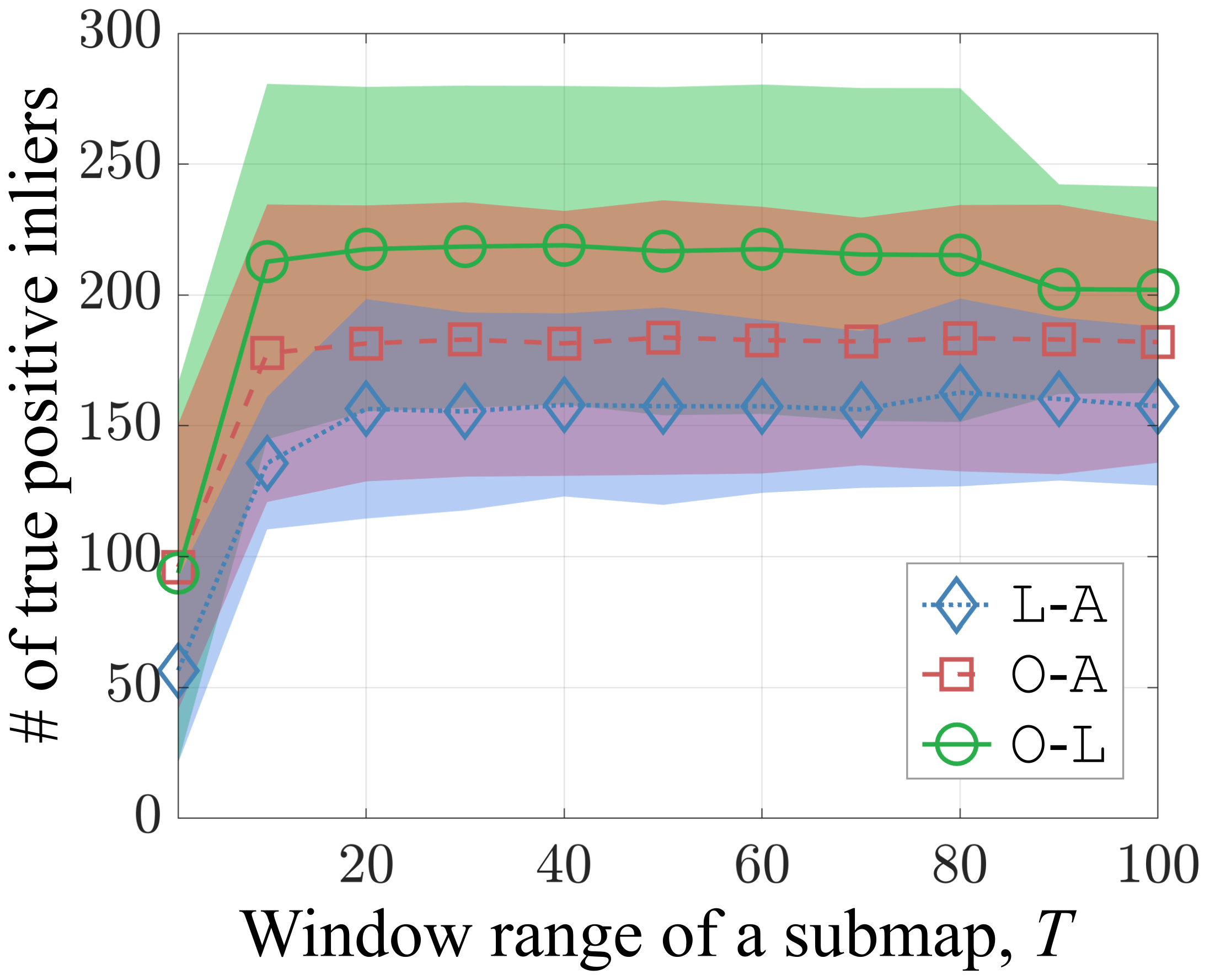}
		\caption{}
		\label{fig:tmse_inlier}
	\end{subfigure}
	\begin{subfigure}[b]{.24\textwidth}
		\centering
		\includegraphics[width=\textwidth]{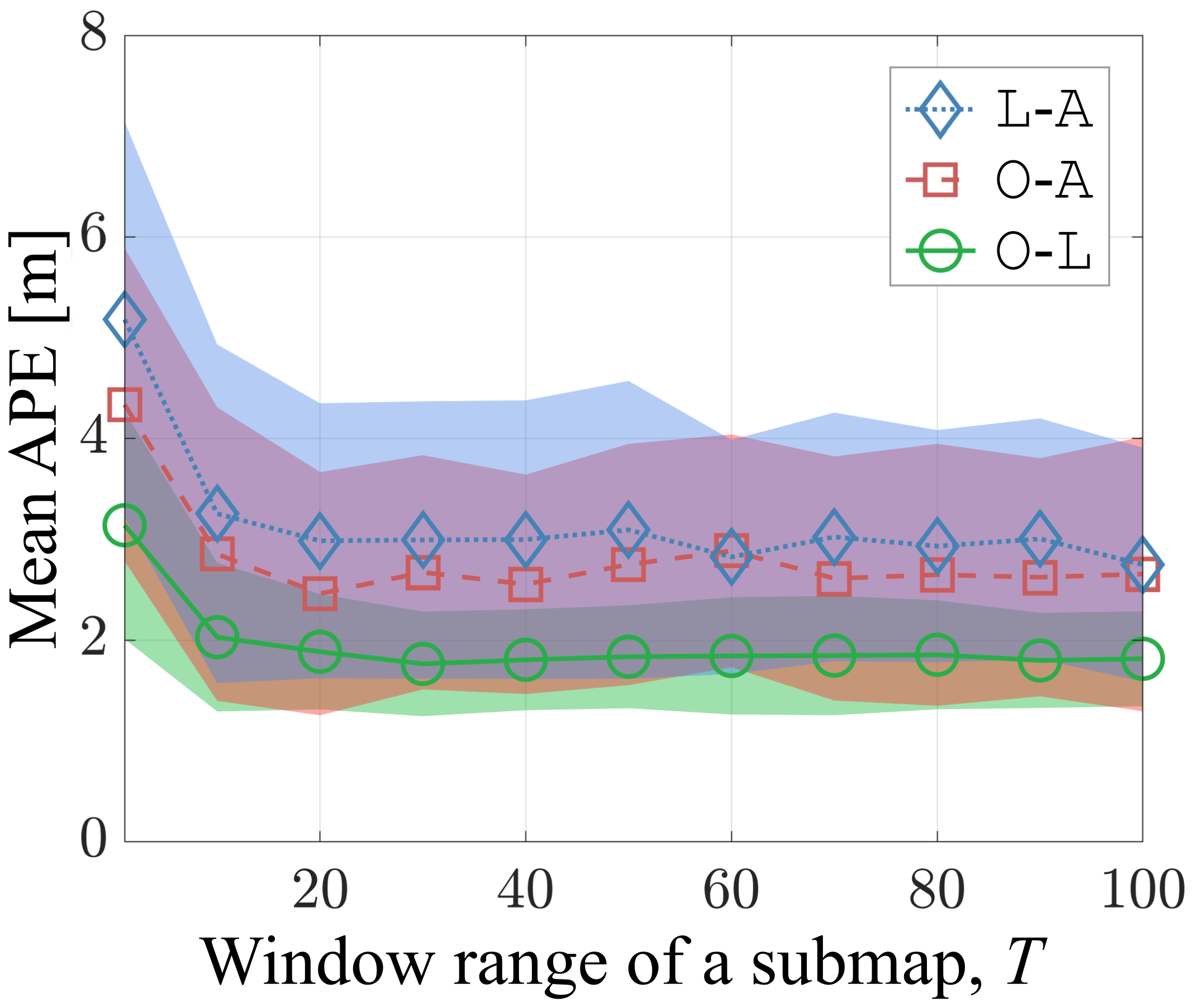}
		\caption{}
		\label{fig:t_ape}
	\end{subfigure}
	\caption{(a)~The average number of true positive inliers and its standard deviation, and (b)~the average mean absolute pose error~(APE) and its standard deviation of multi-session SLAM with varying window range of a submap, $T$, in the HeLiPR dataset
(best viewed in color).}
	\label{fig:t_exp}
    \vsfig
\end{figure}

In addition, we conducted an ablation study to investigate the impact of the submap window range $T$ on the MSS performance.
As shown in \figref{fig:t_exp}(a), setting a sufficiently large $T$ significantly increased the number of true positive loops.
However, when $T$ reached 20, the performance began to saturate, and as $T$ increased further, local drift was more likely to affect the quality of the submaps, causing fluctuations in performance~(\figref{fig:t_exp}(b)).
Therefore, we used $T=20$ in our experiments.

\begin{figure}[t!]
    \captionsetup{font=footnotesize}
    \centering
    \begin{subfigure}[b]{.155\textwidth}
        \centering
        \includegraphics[width=\textwidth]{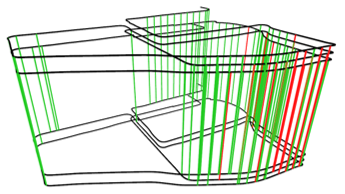}
        \caption{}
    \end{subfigure}
    \begin{subfigure}[b]{.155\textwidth}
        \centering
        \includegraphics[width=\textwidth]{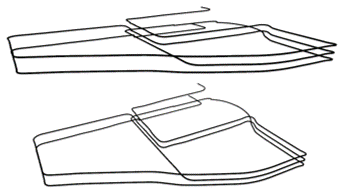}
        \caption{}
    \end{subfigure}
    \begin{subfigure}[b]{.155\textwidth}
        \centering
        \includegraphics[width=\textwidth]{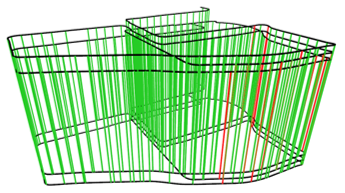}
        \caption{}
    \end{subfigure}
    \caption{(a)-(c)~Visualization of inter-session loop constraints of MSS results on \texttt{DCC05}~(\texttt{O}-\texttt{L}) under different criteria.
        (a)~When mean squared error~(MSE)~$<~30\,\text{m}^2$.
        (b)~When MSE~$<~0.4\,\text{m}^2$.
        (c)~When t-MSE~$<~0.4\,\text{m}^2$, which is a proposed criterion in this paper~(see \secref{sec:scan_to_scan}).
        The green and red lines indicate the true positive and false positive inter-session loop constraints, respectively~(best viewed in color).}
    \label{fig:mse}
    \vsfig
\end{figure}

Furthermore, we also investigated the impact of our t-MSE on the quality of loop constraints.
As shown in~\figref{fig:mse}(a), using MSE allows us to select valid loop constraints, filtering out most false loop constraints.
Nevertheless, undesirable loops (red lines in \figref{fig:mse}(a)) are also considered as valid owing to the partial overlap between two point clouds acquired from heterogeneous LiDAR sensors.
However, if we set the MSE threshold conservatively, all loops are rejected (see \figref{fig:mse}(b)).
In contrast, using our proposed t-MSE showed significant enhancement in loop constraints filtering, enabling more true constraints to remain while reducing the number of false constraints (see \figref{fig:mse}(c)).

\subsection{Runtime Comparison}

Finally, we investigate the runtime of MSS approaches in the initial alignment process to demonstrate its speed and efficiency.
As shown in \tabref{table:time_table}, although MSS is an offline process where runtime is not critical, our algorithm achieved up to approximately 9 times faster than the SC-based approach and up to 5 times faster than the STD-based approach.
This is because LCD-based MSS approaches require the extraction of descriptors for all scans and the complexity of finding inter-session loops increases quadratically (see Desc. ext. columns in \tabref{table:time_table}).

Furthermore, these approaches initialize the inter-session pose difference as an identity matrix that may significantly differ from the actual pose difference.
Consequently, they require more iterations to converge, thereby resulting in significantly longer optimization times.

In contrast, our approach requires only the extraction of FPFH descriptors for the query and central maps with a large voxel size $\vmap$; thus, it only takes a few seconds and the only bottleneck is the voxel sampling of map clouds.
In addition, as listed in \tabref{table:solver_time}, we compared the average solver time with other registration methods, specifically SAC-IA~\cite{rusu2009fpfh}, PCR-99~\cite{lee2024pcr}, and TEASER++~\cite{yang2020teaser}. As a result, Quatro, which was used in our Multi-Mapcher, showed the fastest speed.

Therefore, we demonstrate that our approach is not only robust to heterogeneous LiDAR sensor setups but also performs MSS quickly.

\begin{table}[t!]
    \centering
    \captionsetup{font=footnotesize}
    \caption{Average runtime for initial inter-session alignment in the HeLiPR dataset on Intel(R) Core(TM) i9-13900 CPU~(unit: sec). }
    \setlength{\tabcolsep}{1pt}
    {\scriptsize
        \begin{tabular}{l|cccccccc}
            \toprule \midrule
            \multirow{2}{*}[-0.8em]{Method} & \multicolumn{4}{c}{\texttt{DCC05} (\texttt{O}-\texttt{L})} & \multicolumn{4}{c}{\texttt{KAIST05} (\texttt{O}-\texttt{L})} \\ \cmidrule(lr){2-5} \cmidrule(lr){6-9}
            & \begin{tabular}{@{}c@{}}Voxel \\ sampl.\end{tabular} & \begin{tabular}{@{}c@{}}Desc. \\ ext.\end{tabular} & Optim. & Total & \begin{tabular}{@{}c@{}}Voxel \\ sampl.\end{tabular} & \begin{tabular}{@{}c@{}}Desc. \\ ext.\end{tabular} & Optim. & Total \\ \midrule
            LT-mapper~\cite{kim2022lt} w/ SC~\cite{kim2018scancontext} & \textbf{9.51} & 27.59 & 78.86 & 115.96 & 9.86 & 27.28 & 82.26 & 119.40 \\
            LT-mapper~\cite{kim2022lt} w/ STD~\cite{yuan2023std} & 9.74 & 15.11 & 36.86 & 61.71 & \textbf{9.35} & 16.77 & 35.04 & 61.16 \\
            Multi-Mapcher (Ours) & 10.63 & \textbf{0.46} & \textbf{0.96} & \textbf{12.05} & 11.48 & \textbf{0.68} & \textbf{2.64} & \textbf{14.80} \\
            \midrule \bottomrule
        \end{tabular}
    }
    \label{table:time_table}
\end{table}

\begin{table}[t!]
	\centering
	\captionsetup{font=footnotesize}
	\caption{The average solver time for each point cloud registration approach during inter-session initial alignment on \texttt{KITTI}~Seq.~\texttt{00} and Seq.~\texttt{07}, tested on Intel(R) Core(TM) i9-13900 CPU.}
	{\scriptsize
		\begin{tabular}{lc}
			\toprule \midrule
			Method & Average solver time [ms]   \\ \midrule
			SAC-IA~\cite{rusu2009fpfh} & 1285.29 \\
            PCR-99~\cite{lee2024pcr} & 81.20 \\
			TEASER++~\cite{yang2020teaser} & 8.77 \\
			Quatro~\cite{lim2022single} in Multi-Mapcher (Ours) & \textbf{1.18}  \\
			\midrule \bottomrule
		\end{tabular}
	}
	\label{table:solver_time}
	\vstab
\end{table}

\section{Conclusion}
\label{sec:conclusion}

In this study, we have challenged the existing paradigm that relies heavily on loop detection modules and thus proposed a novel MSS framework, called \textit{Multi-Mapcher}.
By exploiting outlier-robust registration and novel methods to address the differences in the density and FoV of heterogeneous LiDAR sensor setups, our proposed method can precisely perform an inter-session alignment between the query and the central sessions while resolving the dependency on LCD modules.

Despite these encouraging results, there is scope for further research.
Because our Multi-Mapcher is an offline approach, in future work, we will investigate how to extend our approach to online collaborative multi-session SLAM,
exploring how multiple robots can work together in real-time to build a consistent global map.

\section*{Acknowledgments}

We thank Prof. Jaesik Park and Kanghee Lee for conducting fruitful discussions and sharing valuable insights~\cite{lee2022learning} that provided a starting point for our research on map-to-map registration.

\bibliographystyle{URL-IEEEtrans}

\bibliography{URL-bib}

\begin{IEEEbiography}[{\includegraphics[width=1in,height=1.0in]{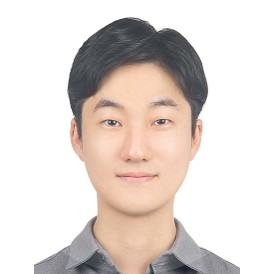}}]{Hyungtae Lim}
	received the B.S. degree in mechanical engineering, and M.S. and Ph.D. degrees in electrical engineering from the Korea Advanced Institute of Science and Technology (KAIST), Daejeon, Republic of Korea, in 2018, 2020, and 2023, respectively.
    He was a postdoctoral fellow in the Information \& Electronics Research Institute, KAIST, Daejeon, Republic of Korea from 2023 to 2024.
    He is currently a postdoctoral associate in the Laboratory for Information \& Decision Systems~(LIDS), Massachusetts Institute of Technology~(MIT), Massachusetts, the U.S.
    His research interests include SLAM (simultaneous localization and mapping), 3D registration, 3D perception, long-term map management, AI, and deep learning.
\end{IEEEbiography}

\begin{IEEEbiography}[{\includegraphics[width=1in,height=1.3in]{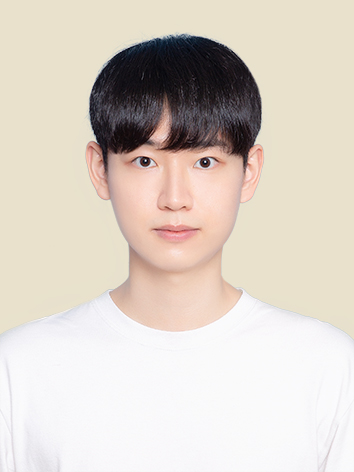}}]{Daebeom Kim}
	received the B.S. degree in automobile and IT convergence from Kookmin University, Seoul, Republic of Korea, in 2023, and the M.S. degree in robotics program from the Korea Advanced Institute of Science and Technology (KAIST), Daejeon, Republic of Korea, in 2025. He is currently pursuing the Ph.D. degree in electrical engineering, KAIST. His research interests include multi-session SLAM, long-term map management, 3D LiDAR perception and registration, robotics, navigation, and autonomous driving.
\end{IEEEbiography}

\begin{IEEEbiography}[{\includegraphics[width=1in,height=1.25in]{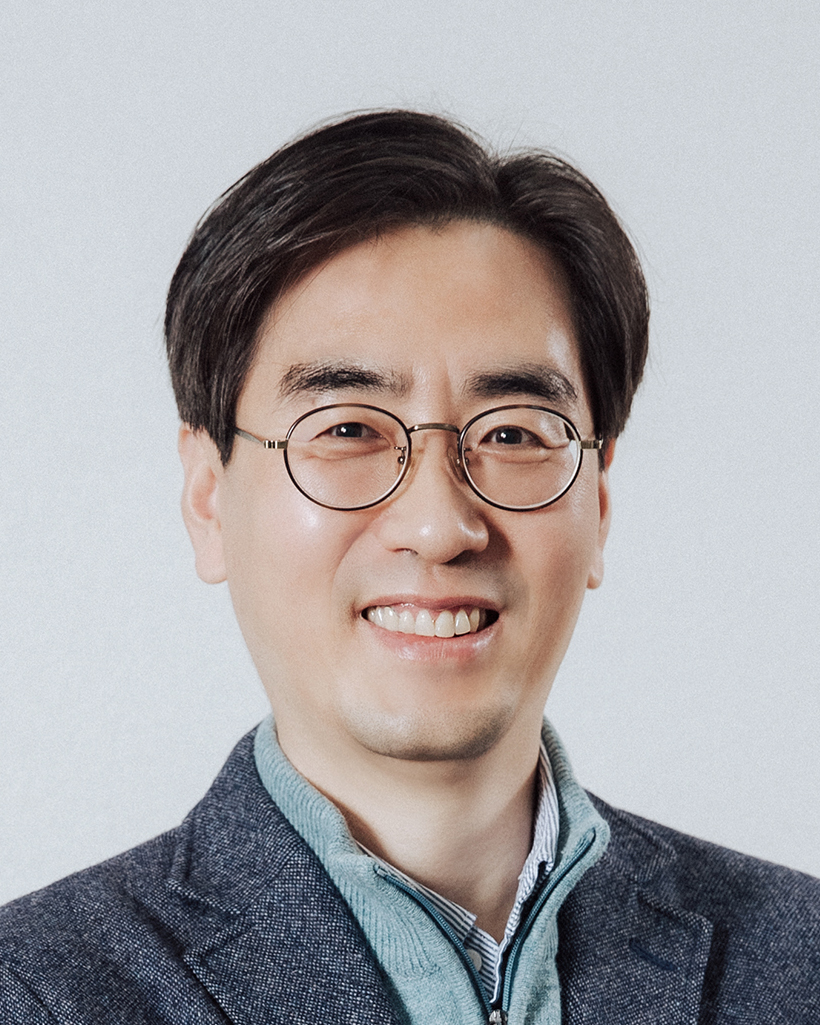}}]{Hyun Myung}
	 received the B.S., M.S., and Ph.D. degrees in electrical engineering from the Korea Advanced Institute of Science and Technology (KAIST), Daejeon, Republic of Korea, in 1992, 1994, and 1998, respectively. He was a Senior Researcher with the Electronics and Telecommunications Research Institute, Daejeon, from 1998 to 2002, a CTO and the Director with the Digital Contents Research Laboratory, Emersys Corporation, Daejeon, from 2002 to 2003, and a Principle Researcher with the Samsung Advanced Institute of Technology, Yongin, Korea, from 2003 to 2008. Since 2008, he has been a Professor with the Department of Civil and Environmental Engineering, KAIST, and he was the Chief of the KAIST Robotics Program. Since 2019, he is a Professor with the School of Electrical Engineering. His current research interests include autonomous robot navigation, SLAM (simultaneous localization and mapping), SHM (structural health monitoring), spatial AI and machine learning, and swarm robots.
\end{IEEEbiography}

\end{document}